\documentclass[lettersize,journal]{IEEEtran}
\usepackage{times}
\usepackage{epsfig}
\usepackage{graphicx,subcaption}
\usepackage{amsmath}
\usepackage{amssymb}
\usepackage{algorithm}
\usepackage{algpseudocode}
\usepackage{color,balance}
\usepackage{subcaption}
\usepackage{tabularx}
\usepackage{breqn}
\usepackage{tablefootnote}
\usepackage{multirow,comment}
\usepackage[TS1,T1]{fontenc}
\usepackage[numbers,sort&compress]{natbib}
\usepackage{lineno}
\usepackage[pagebackref=true,breaklinks=true,colorlinks,bookmarks=false]{hyperref}
\begin{document}

%\title{Towards Scalable and Generalizable Deep Spectral Embedding}
\title{ Learning Structure Aware Deep Spectral Embedding}
%\title{Structural/Structure Preserving Deep Spectral Clustering/Embedding towards Scalable and Generalizable solutions}
%\title{Hybrid Deep Spectral Clustering towards Scalable and Generalizable solutions}
\author{Hira Yaseen,
\and
Arif Mahmood\\  \textcolor{blue}{This is a draft version of the paper accepted in IEEE Transactions on Image Processing, 2023}\\%~\IEEEmembership{Information Technology University,~Pakistan,}

\thanks{H. Yaseen and A Mahmood are with the Department
of Computer Science, Information Technology University (ITU), 346-B, Ferozepur Road, Lahore, Pakistan.
E-mails:  PhDCS17002@itu.edu.pk, arif.mahmood@itu.edu.pk}
}

% The paper headers
%\markboth{IEEE TRANSACTIONS ON IMAGE PROCESSING,~Vol.~,~}%{Shell \MakeLowercase{\textit{et al.}}: A Sample Article Using IEEEtran.cls for IEEE Journals}

%\IEEEpubid{0000--0000/00\$00.00~\copyright~2021 IEEE}
% Remember, if you use this you must call \IEEEpubidadjcol in the second
% column for its text to clear the IEEEpubid mark.

\maketitle

\begin{abstract}
Spectral Embedding (SE) has often been used to map data points from non-linear manifolds to linear subspaces for the purpose of classification and clustering. Despite significant advantages, the subspace structure of data in the original space is not preserved in the embedding space. To address this issue subspace clustering has been proposed by replacing the SE graph affinity with a self-expression matrix. It works well if the data lies in a union of linear subspaces however, the performance may degrade in real-world applications where data often spans non-linear manifolds. To address this problem we propose a novel structure-aware deep spectral embedding by combining a spectral embedding loss and a structure preservation loss. To this end, a deep neural network architecture is proposed that simultaneously encodes both types of information and aims to generate structure-aware spectral embedding. The subspace structure of the input data is encoded by using attention-based self-expression learning. The proposed algorithm is evaluated on six publicly available real-world datasets. The results demonstrate the excellent clustering performance of the proposed algorithm compared to the existing state-of-the-art methods. The proposed algorithm has also exhibited better generalization to unseen data points and it is scalable to larger datasets without requiring significant computational resources.   \end{abstract}


\section{Introduction}
\IEEEPARstart{H}{igh-dimenional} 
\textcolor{blue}{
data often spans  low-dimensional manifolds instead of being uniformly distributed across the ambient space. Recovering these low-dimensional manifolds reduces the computational cost,  memory requirements, and the effect of noise and thus improves the performance of learning, inference, and recognition tasks. Subspace clustering refers to the problem of separating data according to their underlying manifolds. } Subspace clustering algorithms have a wide range of applications in computer vision such as image clustering \cite{chen2020stochastic,seo2019deep,abavisani2020deep,Elhamifar17}, motion segmentation~\cite{xia2017human,sekmen2013subspace}, co-segmentation of 3D bodies~\cite{wu2013unsupervised,Hu12}, DNA sequencing~\cite{wallace2015application,tchagang2014subspace}, omics data clustering~\cite{shi2019multi,ciortan2021optimization}, and gene expression~\cite{wang2019generalized}. %Over the years, many variants of the subspace clustering have been proposed such as Least Squares Regression (LSR)~\cite{Lu12}, Low Rank Representation (LRR)~\cite{Liu14}, Sparse Subspace Clustering (SSC)~\cite{Elhamifar17}, Latent Space SSC (LS3C)~\cite{Patel25}, Kernel SSC~\cite{Patel24}, Block Diagonal Representation~\cite{Lu2019}, Subspace Clustering via Learning and Adaptive Low-rank Graph (SC-LALRG)~\cite{yin2018subspace}, and Stochastic Sparse Subspace Clustering (S$^{3}$COMP-C)~\cite{chen2020stochastic}. 
\textcolor{blue}{Structure of a data is often encoded by using a self-expression matrix which is based on the observation that a data point in a union of subspaces can be efficiently represented by a combination of other data points in the same manifold.}
Over the years, many variants of the subspace clustering have been proposed such as LSR~\cite{Lu12}, LRR~\cite{Liu14}, SSC~\cite{Elhamifar17}, LS3C~\cite{Patel25}, Kernel SSC~\cite{Patel24}, BDR~\cite{Lu2019}, SC-LALRG~\cite{yin2018subspace}, and S$^{3}$COMP-C~\cite{chen2020stochastic}.
In these methods, the self-expression matrix has been computed using conventional optimization techniques. %~\cite{candes2005decoding,tibshirani1996regression}. 
Recently, some algorithms have also used deep neural networks for the computation of self-expression matrices such as DSC~\cite{Ji10}, PSSC~\cite{lv2021pseudo}, NCSC~\cite{zhang2019neural}, S$^{2}$ConvSCN~\cite{zhang2019self}, MLRDSC~\cite{kheirandishfard2020multi}, MLRDSC-DA~\cite{abavisani2020deep}, %DSC-DAG~\cite{yu2020gan}, 
DASC~\cite{zhou2018deep},  SENet~\cite{zhang2021learning}, and ODSC~\cite{valanarasu2021overcomplete}. In all of these algorithms computation of self-expression matrices have been improved by using various constraints and this matrix is then utilized to construct the affinity matrix for spectral clustering. This arrangement works well if the data spans a union of linear subspaces however, the performance may degrade if the underlying space is a nonlinear manifold. Local neighborhood relationships are also ignored which were originally used for the construction of a fully connected graph and thus the strength of spectral embedding is not fully utilized.  
In the current work, we propose to integrate the strength of spectral embedding along with subspace structure preservation using the self-expression matrix. For this purpose, we propose a deep neural network-based architecture that learns embedding by simultaneous minimization of spectral embedding-based loss as well as ensuring self-expression property in the latent space.

The objective function of spectral clustering is to find an embedding of the data points by eigendecomposition of the laplacian matrix encoding pairwise similarities. That data representation is then clustered to assign them to different categories. Despite many advantages, the subspace structure of data in the original space is not preserved in the embedding space. \textcolor{blue}{Structure preservation during data transformation aims to keep a set of embeddings in a common subspace that  shares the same subspace in the ambient space.} 
To address this, many subspace clustering algorithms have been proposed. The assumption of data spanning linear subspaces made by these existing subspace clustering methods often gets violated in many real-world applications. The data may be corrupted by errors or because of missing trajectories, occlusion, shadows, and specularities~\cite{Elhamifar17}. These algorithms compute the self-expression coefficient matrix either using $\ell_1$-norm, $\ell_2$-norm, nuclear norm, or a combination of these norms to preserve data structure. However, local neighborhood information may be lost in these methods because the graph structure and local node connectivity are not included, resulting in sub-optimal clustering performance~\cite{wang2021multi}. 
%Linear subspace clustering methods are not considered reliable to preserve data structure in input data space, which means self-expressiveness does not ensure that close points in high-dimensional space will have similar representations in latent space~\cite{wang2010locality,abdolali2021beyond}. 

In the current work, we explicitly encode both the local and the global input data structures. For encoding the local structure, the supervision of spectral embedding is used to train a deep network preserving local neighborhood graph affinities of data points. For global data structure preservation, the learned embedding is constrained to minimize self-expression loss. The proposed neural architecture is trained in a batch-by-batch fashion by computing both spectral supervision and self-expression in batches. Compared to full data supervision as used by many existing techniques, our proposed approach saves computational time \& memory resources.
%To incorporate connectivity in subspace-structured representation, we build a fully-connected affinity matrix to compute eigen vectors based spectral loss to ensure good connectivity between data points from the same subspace.

The existing self-expressiveness sparse representations may limit connectivity between data points belonging to the same subspace, which may not form a single connected component~\cite{chen2020stochastic}. To handle this issue  $\ell_2$-norm based dense solution has been proposed however it requires the underlying subspaces to be independent~\cite{you2016oracle}. In the current work, we propose a self-attention-based global structure encoding technique. For this purpose, we use two multi-layer fully connected networks including a query net and a key net. These networks are learned by minimizing elastic net-constrained self-expression loss. The learned structure encoding matrix is made sparse by using a nearest-neighbor-based approach removing less probable links in latent space representations. 
%It enables the proposed algorithm to be scalable to larger datasets and generalizable to unseen data points. 

%, using spectral embedding loss and subspace preservation loss, respectively. 
%Our proposed framework jointly learns to emulate both embeddings and produce better results than SOTA methods on different types of datasets such as face clustering, digits clustering, and objects clustering.

%%%%%%%%%%%%%%%%%%%%%%%%%%%%%%%%%%%%%%%%%%%%%%%
%As an added advantage of the current work both generalization and scalability is obtained by using an effective solution based on deep neural networks that learns a set of explicit transformations used by spectral clustering to map input data points into linear subspaces. Thus avoiding computation of affinity matrix as well as computation of eigenvectors of the graph Laplacian matrix for out-of-training data points. For this purpose, an encoder-decoder network is trained to ensure latent space to be close to the spectral embedding by minimizing spectral \& structure-preserving losses. 

Overall the proposed algorithm consists of an end-to-end deep neural architecture that learns structure-aware deep spectral embedding via simultaneous minimization of a Laplacian eigenvector-based loss and a self-attention-based structure encoding loss. As the network gets trained, it iteratively learns the local and global structures of the input data. The proposed algorithm is applied on six publicly available datasets including  EYaleB~\cite{georghiades2001few}, COIL-100~\cite{nene1996columbia}, MNIST~\cite{lecun1998gradient}, ORL~\cite{samaria1994parameterisation}, CIFAR-100~\cite{krizhevsky2009learning}, and ImageNet-10~\cite{chang2017deep},
%GTSRB~\cite{stallkamp2012man}, 
and compared with 51 SOTA methods including deep learning-based methods as well as traditional spectral and subspace clustering techniques. The proposed algorithm has consistently shown improved performance over the compared methods.
The following are the main contributions of the current work:
\begin{itemize}\setlength\itemsep{0em}
    \item We propose to learn non-linear spectral embedding by using the supervision of eigenvectors of the graph Laplacian matrix.
    
    \item The learned embedding is constrained to be structure-aware by using a self-expression-based loss. For this purpose, a self-attention-based structure encoding is exploited.
    
    \item To reduce the complexity both the graph Laplacian and the self-expression matrices are computed batch by batch.  
    
    \item  The proposed deep embedding network is capable of finding effective representations for unseen data points thus enabling better generalization than the existing methods.
   % \item Experiments on six benchmark datasets have demonstrated excellent performance of the proposed algorithm compared to the existing methods.
    %\item In addition to explicit spectral supervision, we also enforce structure preservation which ensures preservation of local and global data structure.
    \end{itemize}
    
%The training process is unsupervised without knowing the class or cluster labels. The proposed algorithm is tested on four publicly available data sets and has shown encouraging performance as compared to the current state of the arts.
The rest of the paper is organized as follows: Section \ref{relatedwork} contains related work, Section \ref{methodology} presents the proposed methodology, and experiments are given in Section \ref{experiments}. Conclusions and future directions follow in Section \ref{conclusion}.
%%%%%%%%%%%%%%%%%%%%%%%%%%%%%%%%%%%%%%%%%%

\section{Related Work}\label{relatedwork}
Due to numerous applications of subspace clustering in computer vision and related fields, several researchers have aimed to improve it in various dimensions such as reducing time complexity~\cite{he2018fast,li2011time}, memory complexity~\cite{shen2016online,li2016scalable}, learning self-expressive coefficients matrix to preserve the linear structure of data~\cite{Ji10}, while some others have intended to generalize it to unseen data~\cite{kang2021structured,shaham32}, and some others have tried to make it scalable~\cite{abdolali2019scalable,kang2021structured,chen2020stochastic}. Chen \textit{et al.}~\cite{chen2020stochastic} proposed random dropout in a self-expressive model to deal with the over-segmentation issues in traditional SSC algorithms. They also used a consensus algorithm to produce a scalable solution over a set of small-scale problems using orthogonal matching pursuit. You \textit{et al.}~\cite{you2018scalable} introduced scalability by dealing with class-imbalanced data using a greedy algorithm that selects an exemplar subset of data using sparse subspace clustering. With the resurgence of deep learning in subspace clustering, many deep network-based methods have also been proposed.

%\textbf{Learning Self expression Based Methods:} 
Ji \textit{et al.}~\cite{Ji10} proposed a convolutional auto-encoder for reconstruction and introduced an additional fully connected self-expressive layer between latent-space data points. Zhang \textit{et al.}~\cite{zhang2019self} further extended the concept of a self-expressive trainable layer by incorporating spectral clustering to compute pseudo labels which are then used to train a classification layer using the latent space representation. Zhang \textit{et al.}~\cite{zhang2019neural} collaboratively used two affinity matrices, one from a trainable self-expressive layer, and another from a binary classifier applying softmax on latent representations, to further improve the self-expression layer. Kheirandishfard \textit{et al.}~\cite{kheirandishfard2020deep} implicitly trained DNN to impose a low-rank constraint on latent space and the self-expressive layer is then used to compute the affinity matrix. Valanarasu \textit{et al.}~\cite{valanarasu2021overcomplete} used over-complete and under-complete auto-encoder networks to get latent representations to input into a trainable self-expressive layer. Lv \textit{et al.}~\cite{lv2021pseudo} used weighted reconstruction loss and a learnable self-expressive layer to compute spectral clustering pseudo labels to be compared with the predictions of a classification layer on top of latent representations. These methods have reported excellent results, however, learning self-expressive coefficients matrix using a full dataset requires high memory complexity. Also if new unseen data points are added, these methods require computing a full self-expressive matrix. Therefore such methods are neither scalable to larger datasets nor generalizable to unseen data points. 

%Chen \textit{et al.}~\cite{chen2018subspace} used DNN to get low-rank latent data representations via a prior calculated low-rank constraint self expressive matrix from input data space. 
Peng \textit{et al.}~\cite{Peng23} computed a prior self-expressive matrix from input data and used it to train an auto-encoder for structure preservation in the latent space.  This method preserves structure however, it requires high computational \& memory complexity to compute self-expressive matrix and train network using a full training dataset. Shaham \textit{et al.}~\cite{shaham32} proposed a DNN to embed input data points into the eigenspace of its associated graph Laplacian matrix. Despite using deep learning, orthogonality in latent space is ensured via QR decomposition instead of using the learning-based framework. Also, input data structure preservation is not explicitly enforced in the latent space.

In contrast to the existing algorithms, we propose to train our network using local neighborhood information using a fully connected graph in addition to the global structure of the data captured by the self-expressive matrix and the reconstruction loss. We use batch-wise training of a fully connected network to produce a subspace-preserving self-expressive matrix at input space enabling scalability to larger datasets and generalization to unseen data points. %To the best of our knowledge, we are the first to train \& embed batch-wise learned self-expressive coefficient matrix to produce structure aware spectral embedding in a DNN. 
%%%%%%%%%%%%%%%%%%%%%%%%%%%%%%%%%%%%%%%%%%%%%%%%%
\section{Proposed Structure Preserving Deep Spectral Clustering}\label{methodology}

In order to implement the proposed solution, we directly train an end-to-end network in a batch-by-batch fashion that learns structure-aware spectral representations of data points. Once this network is trained, new unseen data points can be input into the network, and the corresponding embedding is computed.  %We also train our network on a small subset of data points \& report good clustering accuracy, proposing a scalable solution, as well. 
Since the network is supervised by spectral clustering-based loss, its brief overview is given below.
%We propose to train a separate query-net and key-net using mini batches to train and evaluate an optimal self-expressive coefficient matrix to be used for structure preservation at latent space.
%%%%%%%%%%%%%%%%%%%%%%%%%%%%%%%%%%%%%%%%%%%%
\subsection{Batch-Based Spectral Clustering}
Traditionally spectral clustering has been performed on all data simultaneously however, we propose spectral clustering to be performed on mini-batches and to combine the results using our proposed deep neural network. Compared to the existing methods such an approach would significantly reduce the computational and memory complexity. For this purpose, we divide a given dataset into a large number of random batches, each having $m$ data points. Let  X$=\{x_i\}_{i=1}^m\in \mathcal{R}^{n \times m}$ be a batch data matrix such that $x_i \in \mathcal{R}^{n}$ be a data point spanning $n$ dimensional non-linear manifold. The data batch is mapped to a batch graph $G_b$ having adjacency matrix  $A_b$ $\in \mathcal{R}^{m \times m}$  computed as follows:
\begin{equation}\label{eq:eq1}A_b(i,j)=
    \begin{cases}
    \exp{\left(-\frac{d_{i,j}^{2}}{2\sigma^2}\right)}~~~~~~\text{~~~if~~} i \neq j\\
    ~~~~~0 ~~~~~~~~~~~~~~\text{~~~if~~} i == j
    \end{cases}
\end{equation}
where $d_{i,j}$ is some distance measure between data points $x_i$ and $x_j$ within the same batch, and in our work we consider $d_{i,j}=||x
_i-x_j||_2$.
The graph $G_b$ contains an edge between all nodes $(m_i, m_j)$ representing the local neighborhood, and parameter $\sigma$ controls the width of the neighbourhood~\cite{von2007tutorial}. Laplacian matrix for the graph $G_b$ is computed as $L_b=D_b-A_b$, and $D_b\in \mathcal{R}^{m\times m}$ is a batch based degree matrix defined as:  
\begin{equation}
\label{eq:eq2}
D_b(i,j)=
    \begin{cases}
    \Sigma_{j=1}^{m} A_b(i,j) ~~~~\text{if}~~i==j\\
    ~~~~~0 ~~~~~~\text{~Otherwise}.
    \end{cases}
\end{equation}
The Laplacian matrix $L_b$ is semi-positive definite with at least one zero eigenvalue for a fully connected graph. %The number of zero eigenvalues shows the number of connected components in the graph. The eigenvectors of $L_b$ corresponding to the zero eigenvalues has constant coefficients, while the eigenvectors corresponding to the smallest non-zero eigenvalue can be used to divide a graph into two partitions by just considering the signs of the coefficients. Considering two eigenvectors of $L_b$ corresponding to the two smallest non-zero eigenvalues, $G_b$ can be divided into four partitions ~\cite{Shi2000}. 
Eigenvalue decomposition of $L_b$ is given by: $L_b=U_b\Lambda_b U_b^\top$, where $U_b=\{u_i\}_{i=1}^m$ is a matrix of eigenvectors of $L_b$, such that  $u_i \in \mathcal{R}^{m}$  are arranged in the decreasing order of eigenvalues: $v_{m} \ge v_{m-1} \ge \cdot \cdot \cdot v_{2} \ge v_1$, and $\Lambda_b$ is a matrix having these eigenvalues on the diagonal. For dividing the graph into $k$ partitions, only $k$ eigenvectors corresponding to the minimum non-zero $k$ eigenvalues are considered. \textcolor{blue}{If in a particular batch, the actual number of clusters is less than $k$, even then $k$ eigenvectors are selected. Since the number of clusters is not directly used in the loss function,  a varying  number of clusters across batches has no effect on the training process.} 

Let $U_k=\{u_i\}_{i=1}^k \in \mathcal{R}^{m \times k}$ be the matrix of these eigenvectors. The columns of $U_k^\top$ represent an embedding of original data in a $k$ dimensional space such that the embedding space is linear and therefore any linear clustering algorithm, \textcolor{blue}{such as K-means,} will be able to reveal the groups in the original data. Thus spectral clustering can be considered a projection of data from high-dimensional nonlinear manifolds to low-dimensional linear subspaces. Since we perform spectral clustering batch by batch, to combine all batches to get a unified solution we propose to train a deep neural network to simulate spectral embedding. Such a spectral clustering method would be scalable to significantly larger datasets without incurring computational or memory costs. 
%%%%%%%%%%%%%%%%%%%%%%%
\begin{figure*}
\begin{center}
%\fbox{\rule{0pt}{2in} \rule{.9\linewidth}{0pt}}
\includegraphics[width=1\linewidth]{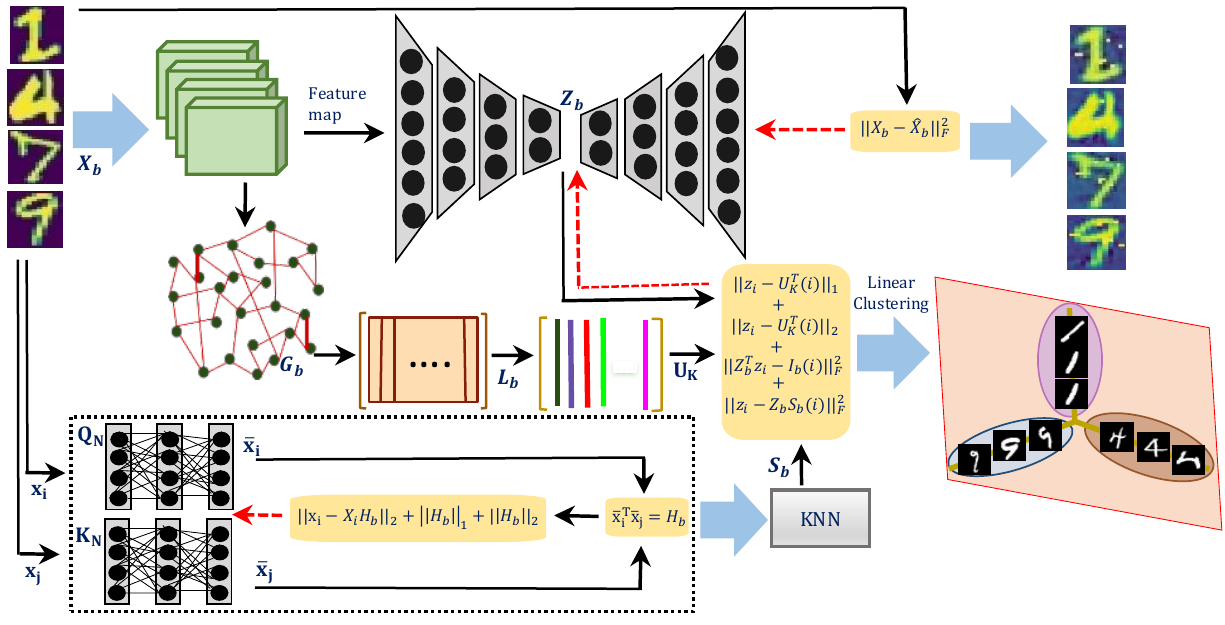}
\end{center}
\vspace{-5mm}
   \caption{The proposed Structure Aware Deep Spectral Embedding (SADSE) network is trained using self-expressiveness and spectral supervision. Batch data matrix $X_b$ is input and the latent space matrix $Z_b$ is constrained to minimize spectral loss and structural losses. The graph $G_b$ is computed batch-wise and is used to compute the Laplacian matrix $L_b$ and eigenvector matrix $U_b$. The latent space of the network is k-dimensional therefore k smallest eigenvectors are selected in $U_k$. $Q_N$ and $K_N$ networks are used to compute $\mathcal{S}_b$.}
\label{fig:model_arch}
\end{figure*}
%%%%%%%%%%%%%%%%%%%%%%%%%%%%%%%%%%%%%%%%%%%%%%%%%%%%%%%%
\subsection{Structure Aware Spectral Embedding}
For larger datasets having millions of data points, the size of the graph and the corresponding Laplacian matrix grows by $O(p^2)$, where $p$ are the total data points. Eigendecomposition on such large matrices incurs a high computational cost. Therefore, traditional spectral clustering lacks scalability to larger datasets ~\cite{Liu2010}.
Similarly, applications with online data arrival require computing an embedding in the low dimensional space without going through the complete process which is also not possible with most of the existing methods. Our proposed solution addresses both of these issues and also incorporates the structure information within the spectral embedding. Fig.~\ref{fig:model_arch} shows all the important steps in our proposed Structure Aware Deep Spectral Embedding (SADSE) algorithm.

A simple strategy to simulate spectral clustering with an auto-encoder is to use the spectral embedding as a direct supervisory signal and train the  network to minimize the error: 
\begin{equation}\label{eq:frob}   \min_{\theta_e} ||U_k^\top-Z_b||^{2}_{F},\end{equation} where $Z_b=\{z_i\}_{i=1}^m \in \mathcal{R}^{k \times m}$ is a k-dimensional latent network embedding such that $k<<n$, and $\theta_e$ are the encoder network parameters. We empirically observe that instead of training only an encoder network, simultaneously training a pair of encoder and decoder networks provides better initialization and improves accuracy.  The encoder and decoder pair referred to as Auto-Encoder is used to project high dimensional data $x_i \in \mathcal{R}^{n}$ to low dimensional latent space $z_i \in \mathcal{R}^{k}$ and then back to the original space.  The encoder may be considered as a nonlinear projector from a higher dimensional to a low dimensional space. Auto-Encoders are often trained to minimize the reconstruction error over a training dataset $X_b$:
\begin{equation}\label{eq:eq4}\mathcal{L}_{R} := \min_{\theta_e, \theta_d} \sum_{i=1}^m||x_i-\widehat {x}_i||_2,\end{equation}
where $\widehat {x}_i$ is the back-projected output of the decoder and $\theta_d$ are the parameters of the decoder network. Reconstruction loss aims to preserve the locality of input data space. We propose to train the deep auto-encoder such that the latent space $Z_b$ be in some sense close to the low dimensional spectral embedding. 

In the current work, instead of \eqref{eq:frob}, we minimize both $\ell_1$ and $\ell_2$ losses between the latent space and spectral embedding as given below:
\vspace{-.1cm}
%\begin{equation}\label{eq:eq5}
\begin{multline}
\mathcal{L}_{S} :=\min_{\theta_e, \theta_d}\sum_{i=1}^m(||x_i-\widehat {x}_i||_2+\lambda_1||z_i-U_{k}^\top(i)||_1+
\\ \lambda_2 ||z_i-U_{k}^\top(i)||_{2}) 
\end{multline}
%\end{equation}
where $U_{k}^\top(i)$ shows a column of $U_{k}^\top$ consisting of the corresponding coefficients from the set of selected $k$ eigenvectors, and $\lambda_1$, and $\lambda_2$ are hyperparameters assigning relative weights to different loss terms, $||.||_{1}$ shows $\ell_1$, and $||.||_{2}$ is $\ell_2$ norm. The $\ell_1$ loss ensures the error is sparse while the $\ell_2$ loss minimizes the distance between the latent space representation $z_i$ and the spectral embedding $U_k^\top(i)$ in the least squares sense. \textcolor{blue}{Due to the convexity enforced by elastic-net regularization, the optimization of the proposed loss is more stable and robust in the presence of noise.}

%Combined reconstruction  and spectral supervision losses are given by  $\mathcal{L}_{R}+\mathcal{L}_{S}$.
%Note that in case of traditional spectral embedding, there is no simple way to back project data to the original space however, using our method one may easily get the back projections using the trained decoder.

Orthogonality on the rows of the matrix $Z_b$ may also be enforced during network training. Since each row of $Z_b$ corresponds to an eigenvector of $L_b$, therefore each row of $Z_b$ is normalized to the unit norm and constrained to be orthogonal to other rows:
%\begin{equation}\label{eq:eq16}
\begin{multline}
\mathcal{L}_{O}:=\min_{\theta_e,\theta_d}\sum_{i=1}^m(||x_i-\widehat {x}_i||_2+\lambda_1||z_i-U_{k}^\top(i)||_1+ \\ \lambda_2||z_i-U_{k}^\top(i)||_{2}+\lambda_3 ||Z_b^\top z_i-I_b(i)||_{F}^{2})
\end{multline}

The spectral embedding does not ensure the input space structure is preserved in the latent space. To this end, we propose a self-expression matrix-based loss function to be simultaneously minimized with the spectral embedding-based loss.   
%%%%%%%%%%%%%%%%%%%%%%%%%%%%%%%%%%%%%%%%%%%%%%%%%
%%%%%%%%%%%%%%%%%%%%%%%%%%%%%%%%%%%%%%%%%%%%%%%%%
In manifold learning, it has been observed that manifold properties may be invariant to some projection spaces~\cite{roweis2000nonlinear}. We aim to find a spectral projection preserving input data structure. For this purpose, a structure-preserving loss simultaneously along with spectral embedding loss is minimized. A pre-computed batch-based self-expressive matrix $\mathcal S_b$ in the input space is utilized for this purpose. The spectral embedding is forced to preserve the input data structure by minimizing the following loss: 
\begin{multline}\label{eq:eq19}
\mathcal{L}_{H}:=\min_{\theta_e,\theta_d}\sum_{i=1}^m(||x_i-\widehat {x}_i||_2+\lambda_1||z_i-U_{k}^\top(i)||_1+ \lambda_2||z_i-\\ U_{k}^\top(i)||_{2}+\lambda_3 ||Z_b^{\top}z_i-I_b(i)||_{F}^{2}+\lambda_4 ||z_i-Z_b \mathcal S_b(i)||_{F}^{2})
\end{multline} where $S_b(i)$ is a column of $S_b$, which is self-expressive representation of data point $x_i$ in the input space.
%%%%%%%%%%%%%%%%%%%%%%%%%%%%%%%%%%%%%%%%%%%%%%%%%

%To preserve global subspace structure of data in the latent space, a self-expressive coefficients matrix from input data matrix $X$ is computed using Lasso for sparse subspace clustering \cite{Peng23, Elhamifar17}. We estimate a batch based self-expressive coefficient matrix $H$ as structure prior by using:
%\begin{equation}\label{eq:eq18}
%    \mathop {min}_{H}||X_b-X_{b}H||_{F}^{2}+\lambda_5||H||_{1} \\~~~ s.t. ~ H(i,i)=0,
%\end{equation}
%where $H(i,i)=0$ denotes diagonal elements of $H$ which correspond to self-reconstruction coefficients. The second term in \eqref{eq:eq18}, \( \ell_1 \)-norm minimization  ensures sparsity of the structure matrix $H$. Table \ref{table:ablation_on_H} shows accuracy results on all datasets when $\phi_{H}$ in \eqref{eq:eq19} is replaced by H.
%%%%%%%%%%%%%%%%%%%%%%%%%%%%%%%%%%%%%%%%%%%%%%%%%0
\subsection{Attention-Based Self-expressive Matrix Learning}
Inspired by the self-attention model in transformer networks, a batch-based self-expressive matrix $\mathcal{H}_b$ is learned using two fully connected learnable networks  $Q_N$ and  $K_N$ \cite{zhang2021learning, vaswani2017attention}. Given a query data point $x_{i}$ which needs to be synthesized using remaining key data points $x_{j}$ in that batch, where $j\neq i$, we forward $x_i$ through $Q_N$: $\Bar{x}_i=Q_N(x_i)\in \mathcal{R}^{t}$ and all $x_j$ through $K_N$: $\Bar{x}_j=K_N(x_j) \in \mathcal{R}^{t}$. 
Attention score between $\Bar {x}_{i}$ and $\Bar {x}_{j}$ is used to get self-expressive coefficients: 
$\mathcal {H}_b(i,j)=\Bar{x}_{i}^\top\Bar{x}_{j}$.  To learn the parameters of $Q_N$ and  $K_N$, the following objective function is minimized: 
\begin{equation}\label{eq:eq22}
\mathop {min}_{\theta_Q,\theta_K} \gamma ||x_i-X_i\mathcal {H}_{b}(i)||_2^2 + \beta||\mathcal{H}_{b}(i)||_1+(1-\beta)||\mathcal{H}_{b}(i)||_2, 
\end{equation}
where $X_i=[x_1, x_2, \cdots x_{i-1},\mathbf{0}, x_{i+1}, \cdots, x_m] $ contains the batch data except $x_i$, $\mathcal {H}_b(i)$ is the $i$-th column of $\mathcal{H}_b$, $\gamma >0$ and $1\ge \beta \ge 0$. In Eq. \eqref{eq:eq22} elastic-net regularizer~\cite{you2016oracle} is used to avoid over-segmentation in $\mathcal {H}_b$. 
Once $\mathcal H_b$ is learned, a sparse binary coefficient matrix $\mathcal S_b$ is computed using KNN algorithm. For each query $x_i$ only the coefficients corresponding to the few nearest neighbors are retained as $1.00$, while the remaining coefficients are suppressed to $0.00$. Thus $\mathcal S_b$ is made sparse enabling only a few nearest neighbors of $x_i$ to contribute. Our choice of batch-wise training of $Q_N$ and  $K_N$, for the computation of $\mathcal{H}_b$, is  scalable to larger datasets and also computationally efficient, and memory requirement is reduced compared to full-scale implementations. %As instead of evaluating $\matchcal H_b$ for all pairs of ($x_{i}, x_{j}$), all $x_i$ are feed forwarded once to obtain representations from $Q_N$ and  $K_N$, and matrix multiplication can then be used in parallel for further computation. Also $Q_N$ and  $K_N$ once trained, can be used to produce $\mathcal H_b$ for unseen data points coming from the same distribution. 

%\subsection{Enforcing Orthogonality on the Latent Space}
%We enforce orthogonality on the rows of the matrix $Z_b$ as a network training step. Since each row of $Z_b$ corresponds to an eigenvector of $L_b$, therefore each row of $Z_b$ is normalized to unit norm and constrained to be orthogonal to other rows  as follows:
%\begin{multline}
%\mathcal{L}_{O}:=\min_{\theta_e,\theta_d}\sum_{i=1}^m(||x_i-\widehat {x}_i||_2+\lambda_1||z_i-U_{k}^\top(i)||_1+ \lambda_2||z_i-U_{k}^\top(i)||_{2} \\ +\lambda_3 ||z_i-Z_b \mathcal S_b(i)||_{F}^{2}+\lambda_4 ||Z_b^{T}z_i-I_b(i)||_{F}^{2})
%\end{multline}

%%%%%%%%%%%%%%%%%%%%%%%%%%%%%%%%%%%%%%%%%%%%%%%%%%%%%%%%%%%%%%%%%%%%%%%%%%
\section{Experimental Evaluations}\label{experiments}
We extensively evaluated the  proposed structure-aware deep spectral embedding (SADSE) algorithm on six publicly available datasets including EYaleB~\cite{georghiades2001few},
Coil-100~\cite{nene1996columbia}, MNIST~\cite{lecun1998gradient}, ORL~\cite{samaria1994parameterisation}, CIFAR-100~\cite{krizhevsky2009learning}, and ImageNet-10~\cite{chang2017deep}, and compare with fifty one existing state-of-the-art approaches including
EDESC~\cite{cai2022efficient},
PSSC~\cite{lv2021pseudo},
SENet~\cite{zhang2021learning},
NCSC \cite{zhang2019neural},
DCFSC \cite{seo2019deep},
S$^{3}$COMP-C~\cite{chen2020stochastic},
ODSC \cite{valanarasu2021overcomplete},
SR-SSC \cite{abdolali2019scalable},
SSCOMP~\cite{you2016scalable}, 
EnSC-ORGEN~\cite{you2016oracle},
DSC~\cite{Ji10},
%DBC~\cite{li2018discriminatively},
DEPICT~\cite{ghasedi2017deep},
Struct-AE~\cite{Peng23}, DASC \cite{zhou2018deep}, 
S$^{2}$Conv-SCN \cite{zhang2019self}, 
MLRDSC \cite{kheirandishfard2020multi}, 
MLRDSC-DA \cite{abavisani2020deep},
S$^{5}$C \cite{matsushima2019selective}, SC-LALRG \cite{yin2018subspace}, KCRSC~\cite{wang2018co}, SpecNet \cite{shaham32}, ACC$\_$CN \cite{li2020autoencoder}, DLRSC \cite{kheirandishfard2020deep}, RGRL-L2 \cite{kang2020relation}, MESC-Net\cite{peng2021maximum}, RED-SC \cite{yang2020residual}, %DSC-DAG \cite{yu2020gan},
RCFE \cite{li2018rank}, FTRR \cite{ma2020towards}, Cluster-GAN~\cite{ghasedi2019balanced}, SSRSC~\cite{xu2019scaled}, S$^{2}$ESC~\cite{zhu2021self}, DSC-DL\cite{huang2020deep}, DAE \cite{vincent2010stacked}, IDEC~\cite{guo2017improved}, DCGAN \cite{radford2015unsupervised}, DeCNN \cite{zeiler2010deconvolutional}, VAE \cite{kingma2013auto}, ADC \cite{haeusser2019associative}, AE \cite{bengio2006greedy}, DEC \cite{xie2016unsupervised}, DAC \cite{chang2017deep}, IIC \cite{ji2019invariant}, DCCM \cite{wu2019deep}, PICA \cite{huang2020deep}, CC \cite{li2021contrastive}, SPICE \cite{niu2022spice}, JULE \cite{yang2016joint}, DDC \cite{chang2019deep}, SCAN \cite{van2020scan}, PCL \cite{li2020prototypical}, and TCL \cite{li2022twin}.

For all datasets, we experimented with two settings including full data to be used as train and test, SADSE$_{F}$, and using an unseen 20\% test data, referred to as SADSE$_{T}$. The proposed approach is compared with SOTA using the measures used by the original authors including classification accuracy (Acc.) and normalized mutual information (NMI). A detailed ablation study is also performed to show the effectiveness of each proposed component. 
%%%%%%%%%%%%%%%%%%%%%%%%%%%%%%%%%%%%%%%%%%%%%%%%%
\subsection{Experimental Settings}
In all of our experiments, we used a four-layer encoder-decoder network  with `tanh' activation function. 
%Weights and biases are initialized using 'glorot\_uniform' and zeros respectively. 
%PCA is used to reduce the input dimensions for each dataset. %to 784 in case of EyaleB, 3000 for coil-100, 2000 for MNIST, and 400 for ORL. 
%The size of latent space $\textbf{z}_i$ is set to 484 for eYaleb, 390 for ORL, 710 for coil-100, and 490 for mnist. The batch size is 486 for EYaleB, 200 for ORL, 720 for coil-100, and 500 for mnist. 
First of all, the full network is trained for 100 epochs using only the reconstruction loss then the encoder network is trained using the remaining proposed losses with Adadalta \cite{Zeiler2012ADADELTAAA} optimizer and a learning rate of $1e^{-3}$. For all experiments, the encoder network is further trained for 1000 epochs. The hyperparameters in Eq. \eqref{eq:eq19} are empirically set to $\lambda_1$ = $\lambda_3$ =0.002, $\lambda_2=\lambda_4$ = 0.02 in all experiments.
Each of  $Q_N$ and $K_N$ is an FC network with three layers beyond the input layer of size \{1024,1024,1024\} and is trained batch by batch. KNN algorithm is then used with 3-nearest neighbors for all datasets. Adam optimizer with a learning rate of $1e^{-3}$, $\beta=0.9$, and $\gamma=200$ are used for the training of $Q_N$ and $K_N$ in all experiments. \textcolor{blue}{As a linear clustering method, K-means is employed.}
%%%%%%%%%%%%%%%%%%%%%%%%%%%%%%%%%%%%%%%%%%%%%%%%%
\subsection{Evaluations on Different Datasets}
\begin{figure}
    \centering
    \begin{subfigure}[b]{0.13\textwidth}
        \fbox{\includegraphics[width=0.9\textwidth]{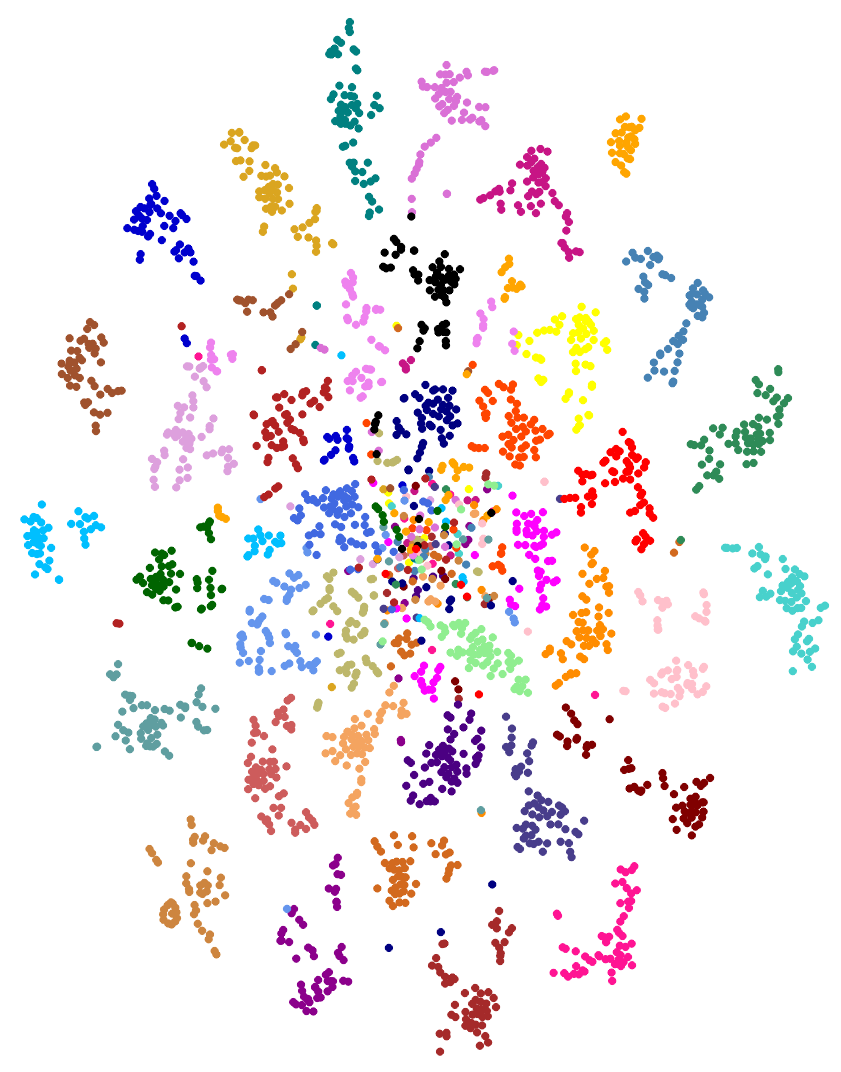}}
        \caption{DLRSC}
        \label{fig:structeyaleb}
    \end{subfigure}
    ~
        \begin{subfigure}[b]{0.132\textwidth}
        \fbox{\includegraphics[width=0.9\textwidth]{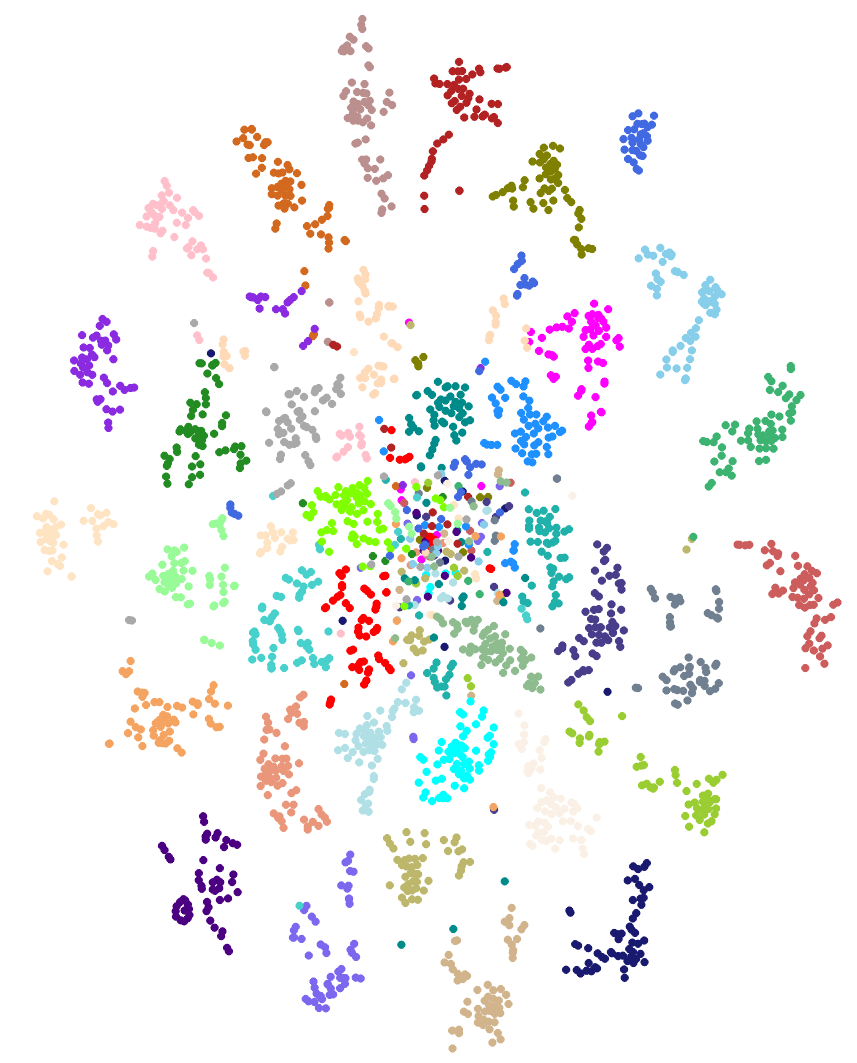}}
        \caption{DSC}
        \label{fig:dscyaleb}
    \end{subfigure}
    ~
     \begin{subfigure}[b]{0.13\textwidth}
        \fbox{\includegraphics[width=0.9\textwidth]{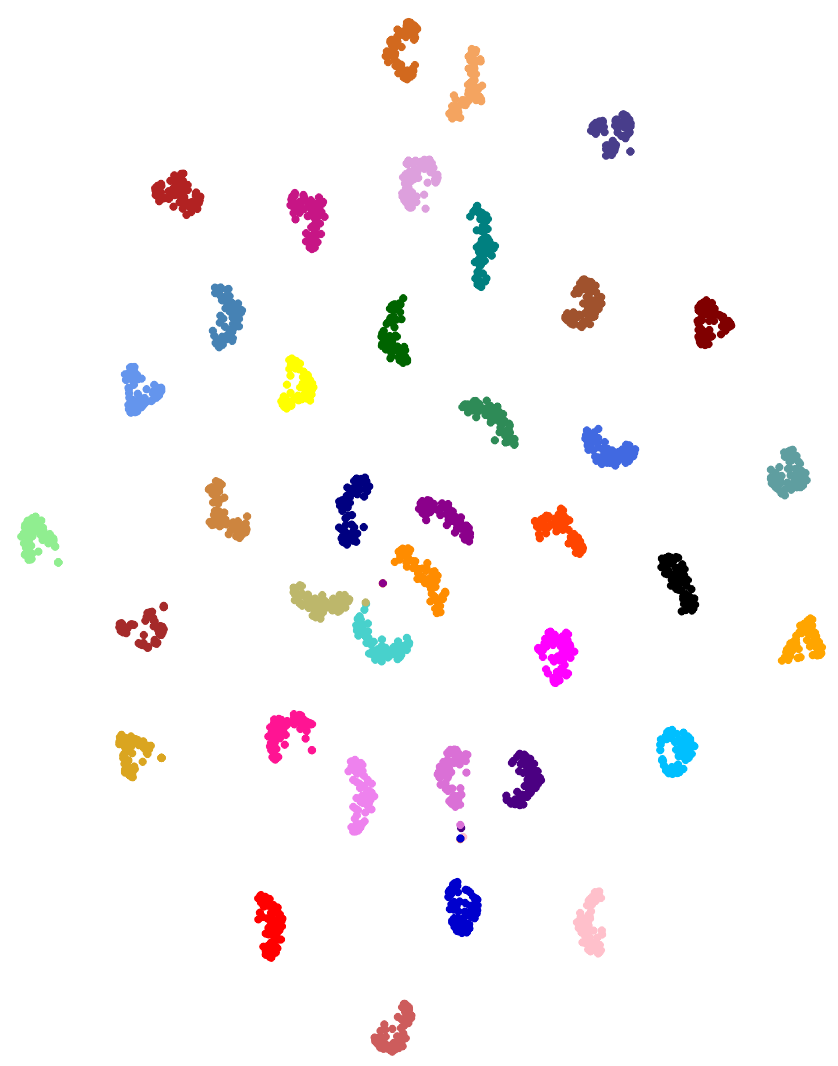}}
        \caption{SADSE$_F$}
        \label{fig:sdscyale}
    \end{subfigure}
    ~
\caption{Visual cluster compactness comparison of the proposed SADSE$_F$ algorithm with SOTA methods using t-SNE on EYaleB dataset}
\label{fig:yaletsne}
\end{figure}

\textbf{EYaleB dataset} contains 64 images having size 192$\times$168 of each of the 38 subjects under 9 different illumination conditions~\cite{lee2005acquiring}. Following the other SOTA methods, we consider only 2432 frontal face images which are then randomly split into 1946/486 train/test splits in SADSE$_{T}$.  Deep features are extracted from the second last layer of Densenet-201 and then PCA is used to reduce dimensions to 784. A batch size of 486 is used. The proposed SADSE$_F$ algorithm has obtained the best accuracy of 99.95\%, outperforming the compared methods as shown in Table~\ref{table:yalestruct}. 
Fig. \ref{fig:yaletsne} shows the visual comparison of the proposed SADSE algorithm with compared methods on this dataset. All 2432 images are plotted using the t-SNE algorithm by assigning each cluster a different color. The clusters obtained by the proposed SADSE algorithm are more compact than the compared methods. 

%%%%%%%%%%%%%%%%%%%%%%%%%%%%%%%%%%%%%%%%%%%%%%%%%%%%%%%%%%%%%%%%%%%
\begin{table}
\begin{center}
\setlength\tabcolsep{7pt}
\begin{tabular}{|l|c|c|c|c|c|}
\hline %inserts double horizontal lines
{} &  \multicolumn{2}{c|}{EYALEB} & \multicolumn{2}{c|}{MNIST}\\
\hline
{\bf Methods} & {\bf Acc.} & {\bf NMI}& {\bf Acc.}&{\bf NMI}\\
\hline
S$^{5}$C \cite{matsushima2019selective}&60.70&-&59.60&-\\
SSCOMP \cite{you2016scalable}&77.59&83.25&-&-\\
SC-LALRG \cite{yin2018subspace}&79.66&84.52&78.20&76.01\\
KCRSC~\cite{wang2018co}&81.40&88.10&64.70&64.30\\
S$^{3}$COMP-C \cite{chen2020stochastic}&87.41&86.32&96.32&-\\
FTRR \cite{ma2020towards}&-&-&70.70&66.72\\
PSSC$_{l}$~\cite{lv2021pseudo}&-&-&78.50&72.76\\
PSSC~\cite{lv2021pseudo}&-&-&84.30&76.76\\
DCFSC \cite{seo2019deep}&93.87&-&-&-\\
%PARTYs \cite{peng2016deep}&94.50&-\\
Struct-AE \cite{Peng23}&94.70&-&65.70&68.98\\
DEC~\cite{xie2016unsupervised}&-&-&84.30&-\\
IDEC~\cite{guo2017improved}&-&-&88.06&86.72\\
SR-SSC \cite{abdolali2019scalable}&-&-&91.09&93.06\\
EDESC \cite{cai2022efficient}&-&-&91.30&86.20\\
EnSC-ORGEN~\cite{you2016oracle}&-&-&93.79&-\\
NCSC \cite{zhang2019neural}&-&-&94.09&86.12\\
DSC-Net-L1 \cite{Ji10}&96.67&-&-&-\\
ACC$\_$CN \cite{li2020autoencoder}&97.31&99.34&78.60&74.21\\
DSC-Net-L2 \cite{Ji10}&97.33&-&-&-\\
DLRSC \cite{kheirandishfard2020deep}&97.53&-&-&-\\
RGRL-L2 \cite{kang2020relation}&97.53&96.61&81.40&75.52\\
ODSC \cite{valanarasu2021overcomplete}&97.78&-&81.20&-\\
MESC-Net\cite{peng2021maximum}&98.03&97.27&81.11&82.26\\
Cluster-GAN~\cite{ghasedi2019balanced}&-&-& 96.40&92.10\\
DEPICT~\cite{ghasedi2017deep}&-&-&96.50&91.70\\
SENet~\cite{zhang2021learning}&-&-&96.80&91.80\\
SpecNet \cite{shaham32}&-&-&97.10&92.40\\
S$^{2}$Conv-SCN-L2 \cite{zhang2019self}&98.44&-&-&-\\
S$^{2}$Conv-SCN-L1 \cite{zhang2019self}&98.48&-&-&-\\
RED-SC \cite{yang2020residual}&98.52&-&74.34&73.16\\
DASC \cite{zhou2018deep}&98.56&98.01&80.40&78.00\\
MLRDSC \cite{kheirandishfard2020multi}&98.64&-&-&-\\
DSC-DL\cite{huang2020deep}&98.90&97.40&81.20&76.10\\
%DSC-DAG \cite{yu2020gan}&99.12&98.54\\
MLRDSC-DA \cite{abavisani2020deep}&99.18&-&-&-\\
\hline
SADSE$_{F}$&{\bf 99.95}&{\bf 99.95}& \bf {97.35} &\bf{92.81}\\
\hline
\end{tabular}
\vspace{-2mm}
\end{center} 
\caption{Comparison of the proposed SADSE$_F$ algorithm with existing SOTA on the EYaleB and MNIST datasets using full data as train and test.}
\label{table:yalestruct} 
\end{table}
%%%%%%%%%%%%%%%%%%%%%%%%%%%%%%%%%%%%%%%%%%%
%\begin{figure}[!h]
%\begin{center}
%\includegraphics[width=0.85\linewidth]{latex/figures/coil100accuracy.pdf}
%\vspace{-2mm}
%\caption{Stability comparison of different SOTA methods during training on Coil-100.}
%\label{fig:itercoil}
%\end{center}
%\end{figure}
%%%%%%%%%%%%%%%%%%%%%%%%%%%%%%%%%%%%%%%%%%%%%

\begin{figure}
    \centering
    \begin{subfigure}[b]{0.128\textwidth}
        \fbox{\includegraphics[width=0.9\textwidth]{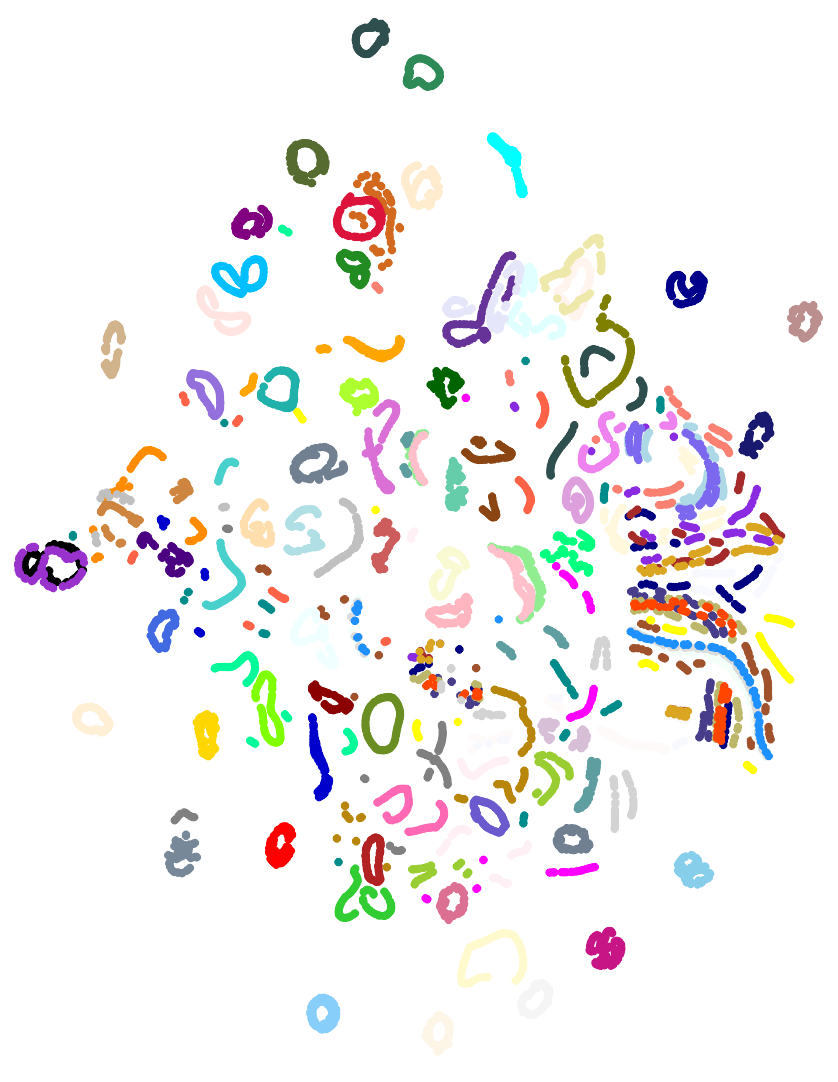}}
        \caption{DLRSC}
        \label{fig:dlrsccoil100}
    \end{subfigure}
    ~
     \begin{subfigure}[b]{0.139\textwidth}
        \fbox{\includegraphics[width=0.9\textwidth]{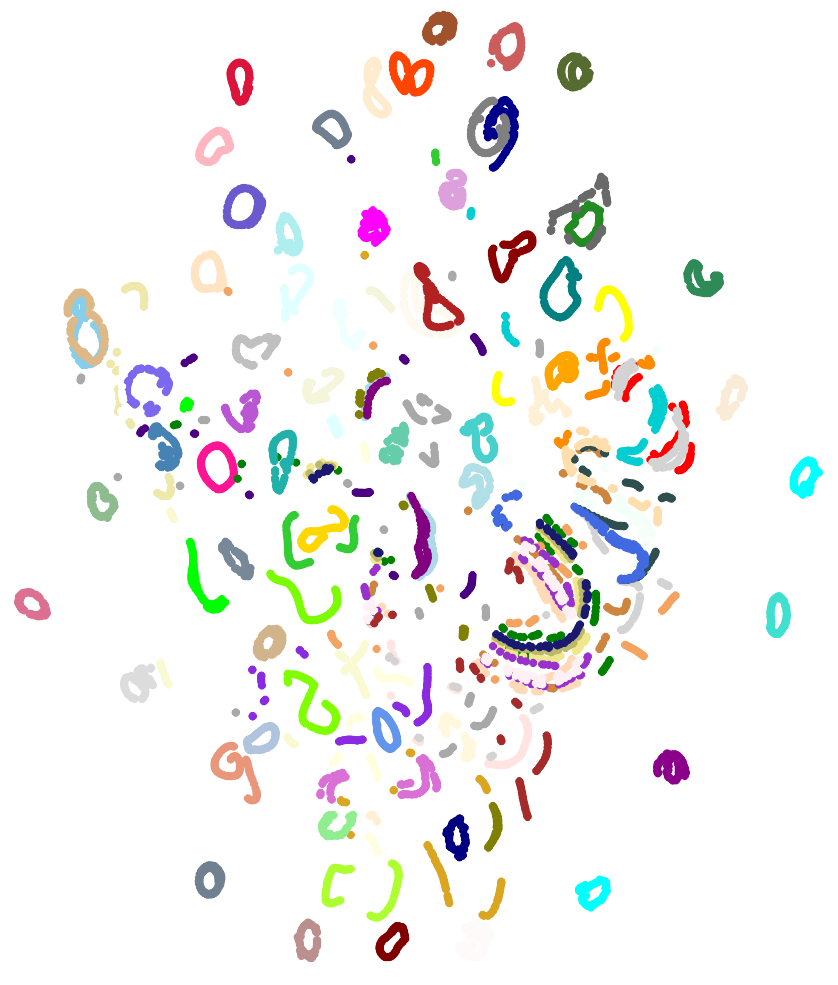}}
        \caption{DSC}
        \label{fig:dsccoil100}
    \end{subfigure}
    ~
     \begin{subfigure}[b]{0.13\textwidth}
        \fbox{\includegraphics[width=0.9\textwidth]{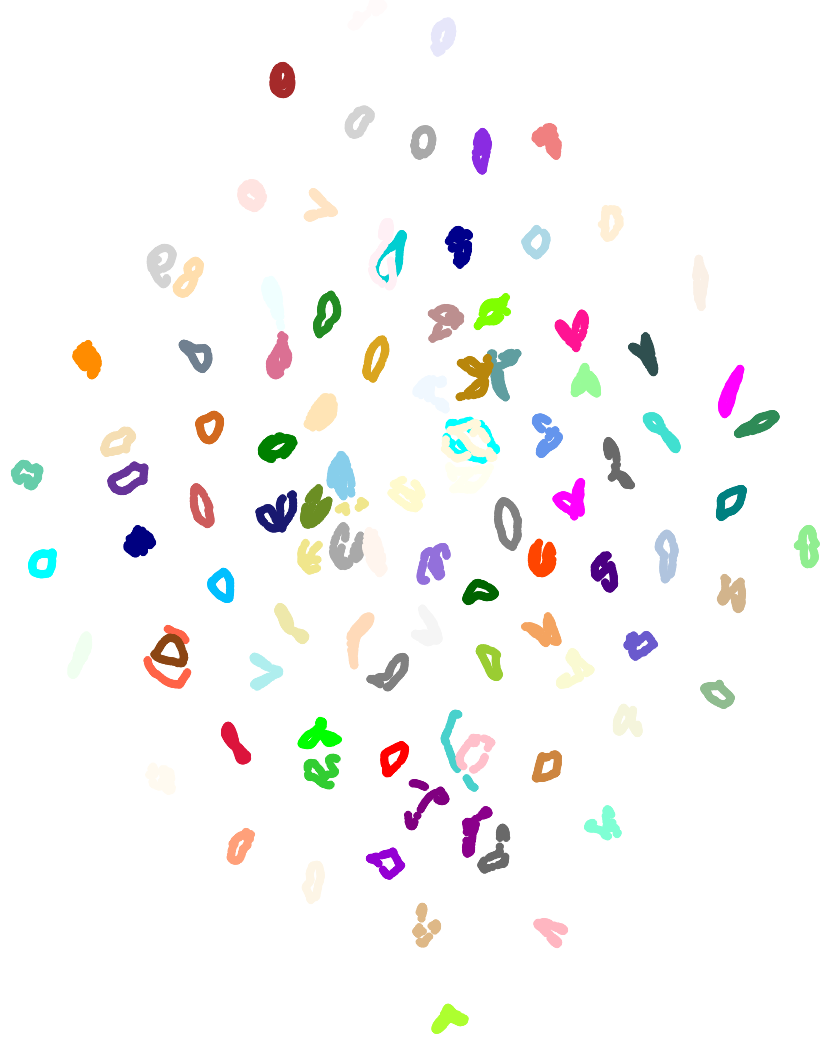}}
        \caption{SADSE$_F$}
        \label{fig:sdsccoil100}
    \end{subfigure}
    ~
\vspace{-2mm}
\caption{Visual cluster compactness comparison of the proposed SADSE$_{F}$  with SOTA methods using t-SNE on Coil-100 dataset.}
\label{fig:coil100tsne}
\end{figure}

\textbf{Coil-100 dataset} has 7200 gray-scale images having size 128$\times$128 of 100 different objects taken at pose intervals of 5 degrees. Deep features are extracted from the second last layer of DenseNet-201 and then PCA is used to reduce dimensionality to 3000. We used a batch size of 720, and a random train/test split of 5760/1440 in SADSE$_{T}$. The proposed SADSE$_{F}$ has obtained an accuracy of 84.95\% outperforming SOTA methods as shown in Table~\ref{table:coil100}. Fig. \ref{fig:coil100tsne} shows a visual comparison of cluster compactness for different algorithms.

\begin{table}
\begin{center}
\setlength\tabcolsep{7pt}
\begin{tabular}{|l|c|c|c|}
\hline
{\bf Methods} & {\bf Acc.}&{\bf NMI}\\
\hline
S$^{5}$C \cite{matsushima2019selective}&54.10&-\\
DSC-Net-L1\cite{Ji10}&66.38&-\\
DSC-Net-L2\cite{Ji10}&69.04&-\\
EnSC-ORGEN ~\cite{you2016oracle}&69.24&-\\
DLRSC \cite{kheirandishfard2020deep}&71.86&-\\
%DSC-DAG \cite{yu2020gan}&71.87&74.37\\
MESC-Net\cite{peng2021maximum}&71.88&90.76\\
S$^{2}$Conv-SCN-L2 \cite{zhang2019self}&72.17&-\\
DCFSC \cite{seo2019deep}&72.70&-\\
S$^{2}$Conv-SCN-L1 \cite{zhang2019self}&73.33&-\\
MLRDSC \cite{kheirandishfard2020multi}&76.72&-\\
%DBC \cite{li2018discriminatively}&77.50&90.50\\
S$^{3}$COMP-C \cite{chen2020stochastic}&78.89&-\\
MLRDSC-DA \cite{abavisani2020deep}&79.33&-\\
RCFE \cite{li2018rank}&79.63&{\bf 96.23}\\
\hline
{SADSE$_{F}$}~&~{\bf 84.95}~&93.91\\
\hline
\end{tabular}
\end{center}
\vspace{-2mm}
\caption{Comparison of the proposed SADSE$_{F}$ algorithm with existing SOTA methods on Coil-100 dataset.}
\label{table:coil100} 
\end{table}
%%%%%%%%%%%%%%%%%%%%%%%%%%%%%%%%%%%%%%%%%%%%%%%

\begin{figure}[!h]
\begin{center}
\includegraphics[width=1\linewidth]{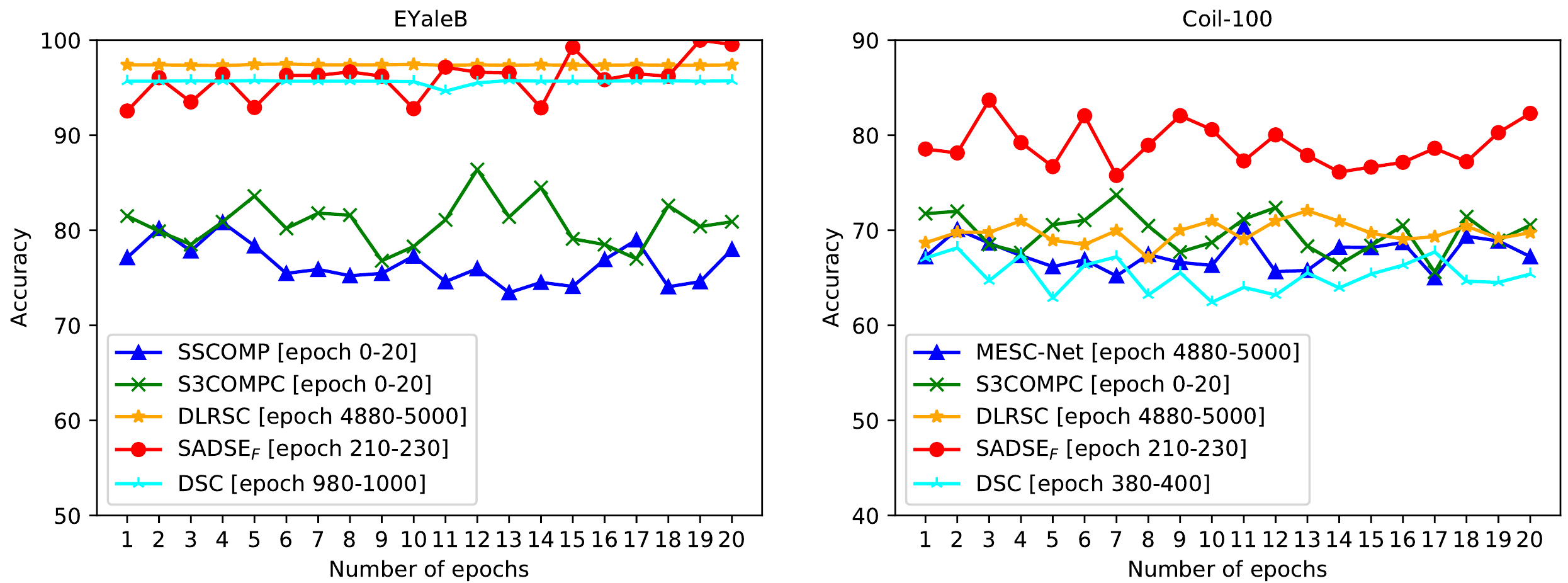}
%\vspace{-2mm}
\caption{Stability comparison of SADSE$_F$ with different SOTA methods during training on EyaleB and Coil-100.}
\label{fig:iter}
\end{center}
\end{figure}

%%%%%%%%%%%%%%%%%%%%%%%%%%%%%%%%%%%%%%%%%%%%%%%%%
\noindent{\textbf{Training stability of SADSE algorithm}:}
In Fig.~\ref{fig:iter}, we compare the training performance stability of different SOTA methods on EYaleB and Coil-100 datasets, respectively. The existing compared methods were trained using a single batch over the full dataset in each epoch while SADSE is trained using 5 batches for EYaleB and 10 batches for Coil-100 in each epoch. Despite batch-based training, the proposed method does not fluctuate much compared to the SOTA methods. The stability of SADSE demonstrates better convergence as the model gets trained. 
%%%%%%%%%%%%%%%%%%%%%%%%%%%%%%%%%%%%%%%%%%

%%%%%%%%%%%%%%%%%%%%%%%%%%%%%%%%%%%%%%%%%%%%%
\textbf{MNIST dataset} consists of 70000 grayscale images of size 28$\times$28 with 10 different classes~\cite{lecun1998gradient}. For this dataset, we computed scattered convolutional features using~\cite{bruna2013invariant} as a pre-processing step, as used by other SOTA ~\cite{chen2020stochastic}. PCA is then used to reduce dimensionality to 2000 and a batch size of 500 is used. The standard 60000/10000 train/test split is used in SADSE$_{T}$. The proposed SADSE$_{F}$ has obtained an accuracy of 97.35\% compared to SOTA methods as shown in Table~\ref{table:yalestruct}. 

\textbf{ORL dataset} contains 400 face images of size $112\times92$ of 40 different subjects, where each subject presents 10 images of different facial expressions under varying light conditions~\cite{samaria1994parameterisation}. The dataset exhibits face images with open or closed eyes, wearing glasses or not, and having a smile or not. A random train/test split of 320/80 is used in SADSE$_T$. Deep features are extracted from the second last layer of Densenet-121 and PCA is used to reduce dimensionality to 400 with a batch size of 400. The proposed SADSE$_{F}$ has obtained an accuracy of 90.75\% outperforming SOTA methods as shown in Table~\ref{table:orl}. 
\begin{table}
\begin{center}
\setlength\tabcolsep{7pt}
\begin{tabular}{|l|c|c|c|}
\hline
{\bf Methods} & {\bf Acc.}&{\bf NMI}\\
\hline
KCRSC~\cite{wang2018co}&72.30&86.30\\
SSRSC~\cite{xu2019scaled}&78.25&-\\
DCFSC \cite{seo2019deep}&85.20&-\\
PSSC$_{l}$~\cite{lv2021pseudo}&85.25&92.58\\
DSC-Net-L1\cite{Ji10}&85.75&-\\
DSC-Net-L2\cite{Ji10}&86.00&-\\
RED-SC \cite{yang2020residual}&86.13&91.16\\
PSSC~\cite{lv2021pseudo}&86.75&93.49\\
DASC~\cite{zhou2018deep}&88.25&93.15\\
S$^{2}$Conv-SCN-L2 \cite{zhang2019self}&88.75&-\\
MLRDSC \cite{kheirandishfard2020multi}&88.75&-\\
S$^{2}$ESC~\cite{zhu2021self}&89.00&93.52\\
S$^{2}$Conv-SCN-L1 \cite{zhang2019self}&89.50&-\\
%DSC-DAG \cite{yu2020gan}&{\bf 90.11}&{\bf 94.78}\\
%%%S$^{2}$DSC-AG \cite{yu2020gan}&91.09&94.97\\
\hline
{SADSE$_{F}$}~&~\bf {90.75}~& \bf {94.66}\\
\hline
\end{tabular}
\end{center}
\vspace{-2mm}
\caption{Comparison of the proposed SADSE$_F$ algorithm with existing SOTA methods on ORL dataset.}
\label{table:orl} 
\end{table}

%%%%%%%%%%%%%%%%%%%%%%%%%%%%%%%%%%%%%%%%
\begin{table}[b]

\begin{center}
\setlength\tabcolsep{7pt}
%\resizebox{\columnwidth}{!}{%
\begin{tabular}{|c|c|c|c|c|c|c|c|}
\hline 
\bf{Datasets}&\bf{Methods}&\bf{Acc.}&\bf{NMI}&\bf{F1 score}&\bf{Precision}\\
\hline
\multirow{2}{*}{EyaleB}
&SADSE$_{F}$&99.95&99.95&99.95&99.95\\
\cline{2-6}
&SADSE$_{T}$&99.95&99.95&99.95&99.95\\
\hline
\hline
\multirow{2}{*}{Coil-100}
&SADSE$_{F}$&84.95 &93.91&84.95&84.57\\
\cline{2-6}
&SADSE$_{T}$&86.04&94.17&86.04&86.26\\
\hline
\hline
\multirow{2}{*}{MNIST}
&SADSE$_{F}$&97.35&92.81&97.35&97.35\\
\cline{2-6}
&SADSE$_{T}$&97.43&93.23&97.43&97.43\\
\hline
\hline
\multirow{2}{*}{ORL}
&SADSE$_{F}$&90.75&94.66&90.75&92.61\\
\cline{2-6}
&SADSE$_{T}$&87.50&95.91&87.50&88.75\\
\hline
\hline
\multirow{2}{*}{CIFAR-100}
&SADSE$_{F}$&47.75&45.77&47.75&47.86\\
\cline{2-6}
&SADSE$_{T}$&47.79&46.47&47.79&52.30\\
\hline
\hline
\multirow{2}{*}{ImageNet-10}
&SADSE$_{F}$&91.69&87.53&91.69&91.78\\
\cline{2-6}
&SADSE$_{T}$&90.43&87.33&90.43&90.55\\
\hline
%\multirow{2}{*}{GTSRB}
%&DSE$_{full}$&&&&\\
%\cline{2-6}
%&DSE$_{test}$&&&&\\
%\hline
\end{tabular}
%} 
\vspace{-2mm}
\caption{Performance comparison of the proposed SADSE$_{T}$ algorithm over unseen test split and full dataset used for both training  and testing (SADSE$_{F}$). }
\label{table:genscale} 
\end{center}
\vspace{-5mm}
\end{table}
%%%%%%%%%%%%%%%%%%%%%%%%%%%%%%%%%%%%%%%%%%%%%%
\begin{figure*}[h]
\centering
    \begin{subfigure}{0.33\textwidth}
    \centering
        \includegraphics[width=1\linewidth]{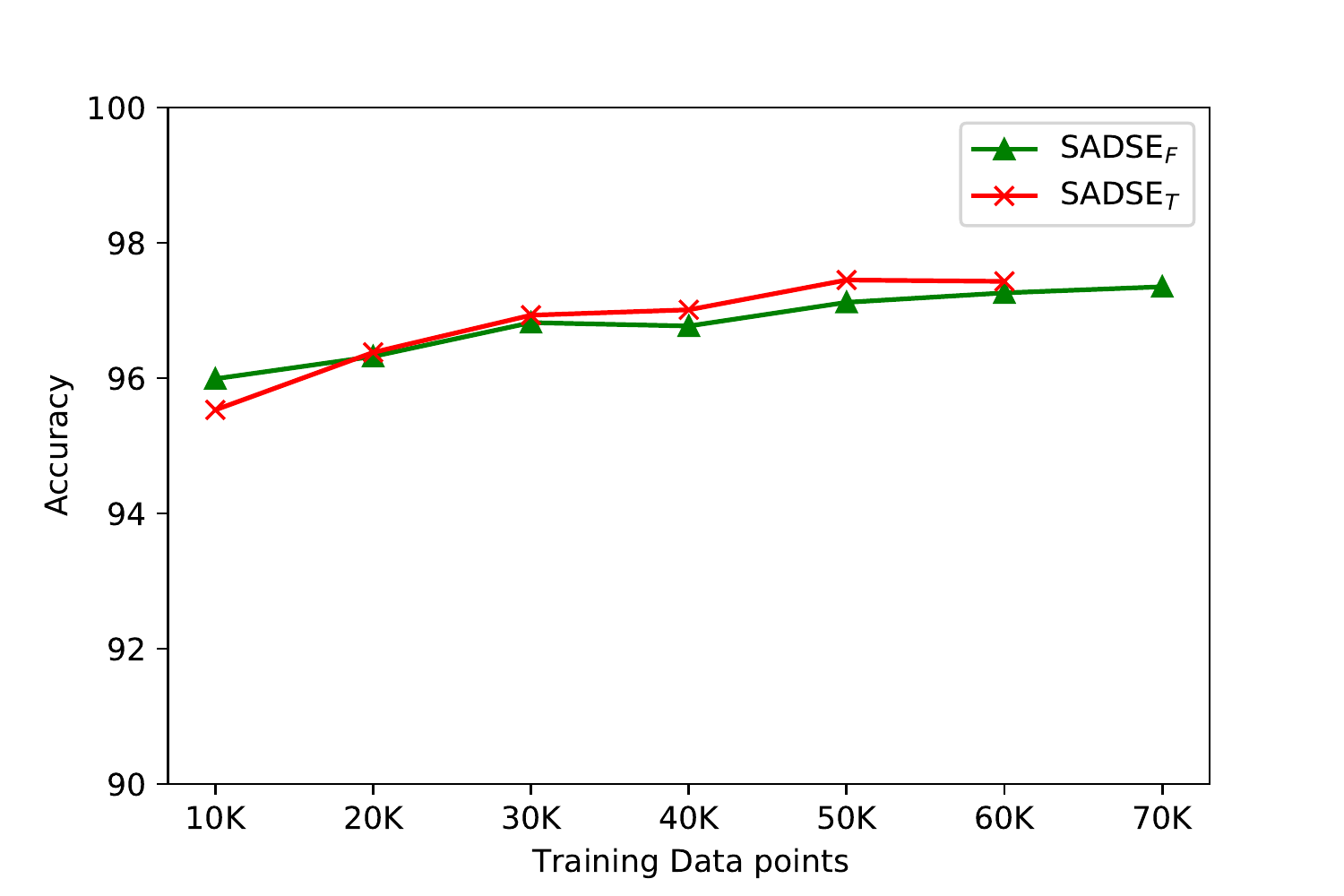}
        \subcaption{}
    \end{subfigure}%
\begin{subfigure}{0.33\textwidth}
\centering
        \includegraphics[width=1\linewidth]{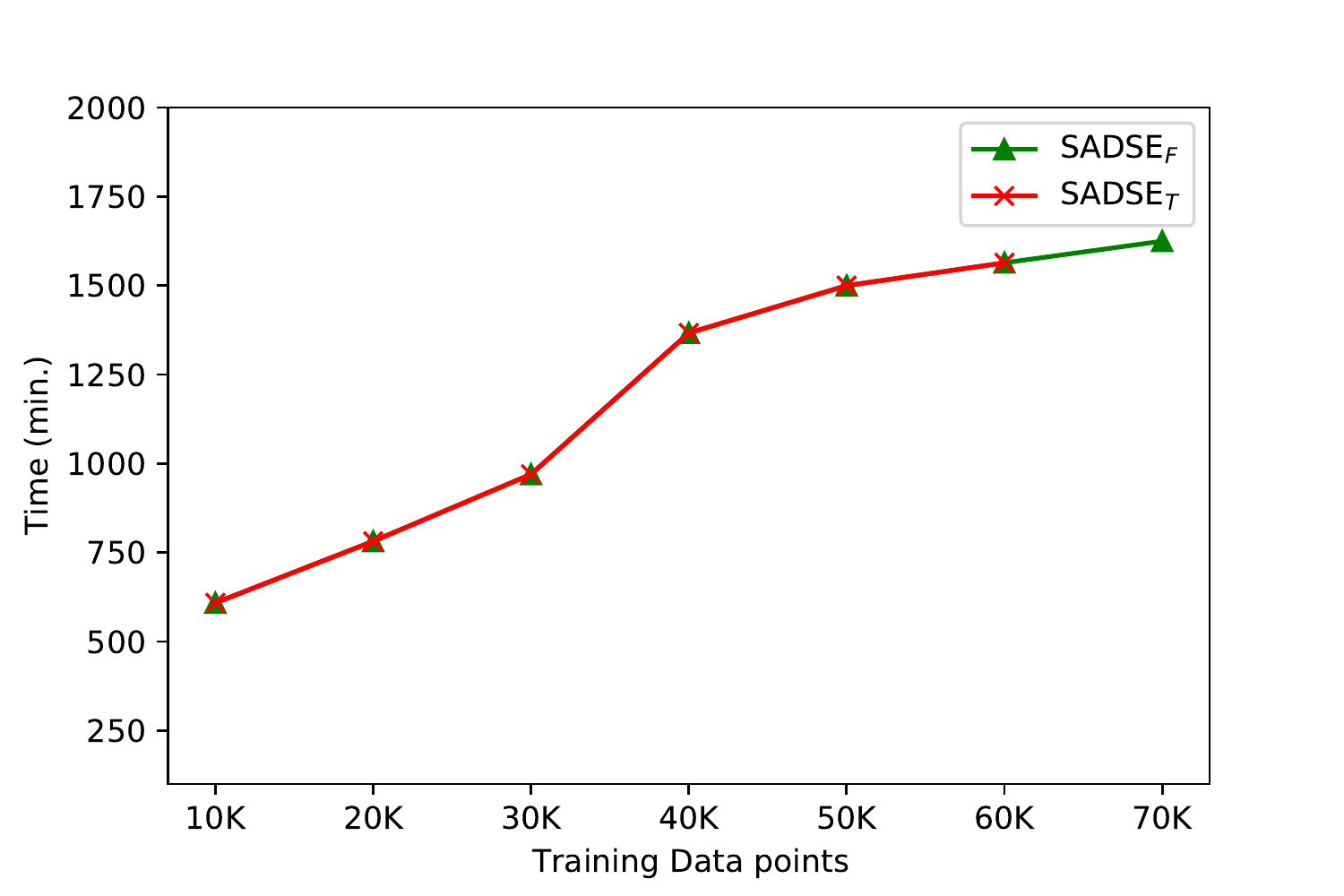}
        \subcaption{}
\end{subfigure}%
\begin{subfigure}{0.33\textwidth}
\centering
        \includegraphics[width=1\linewidth]{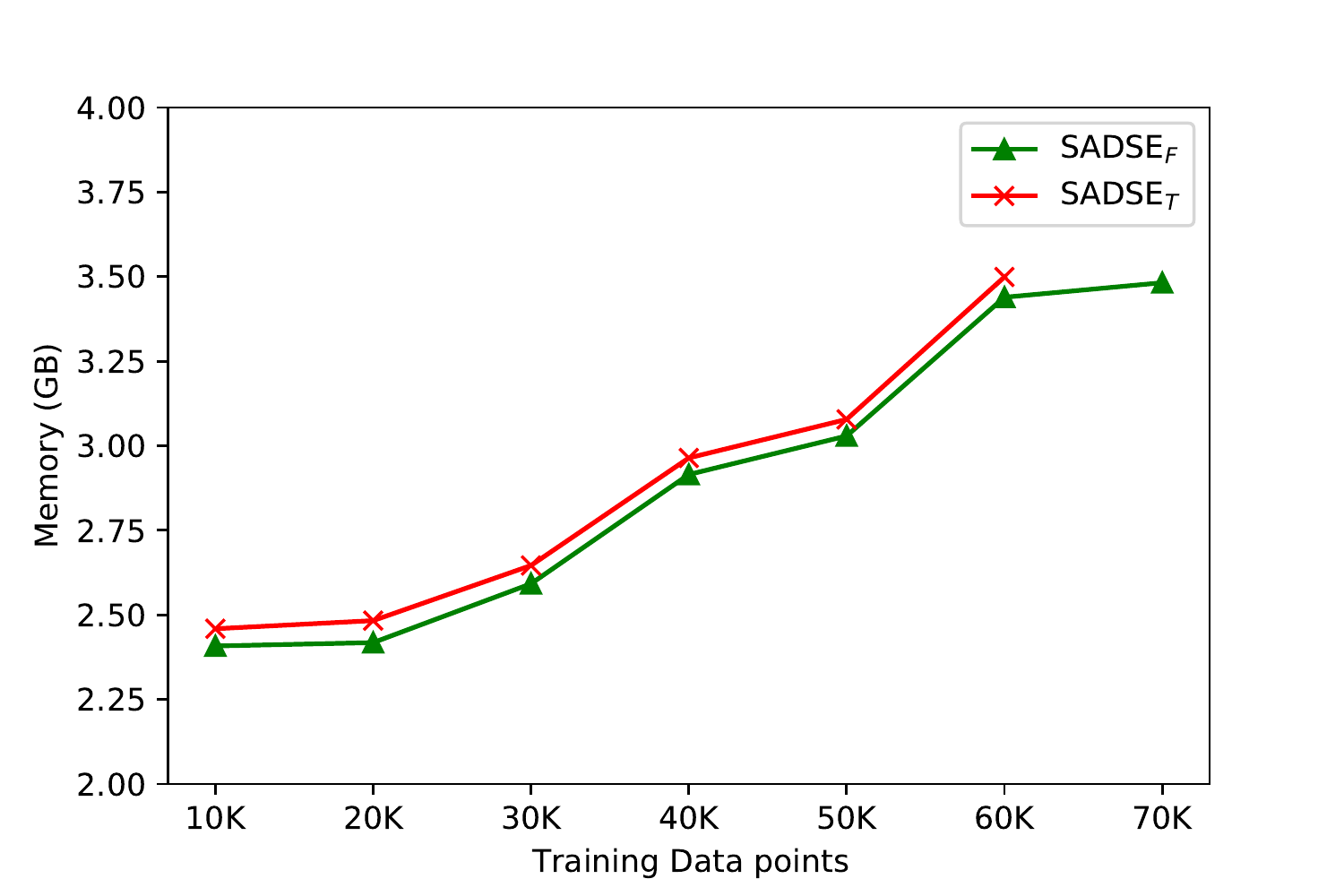}
        \subcaption{}
    \end{subfigure}%
\vspace{-3mm}
\caption{Scalability of the SADSE algorithm on the MNIST dataset is evaluated by observing the variation of accuracy, execution time, and memory consumption, by increasing dataset size. This experiment shows SADSE algorithm is  scalable to larger datasets.}
\label{fig:ablationmnist}
\end{figure*}
%%%%%%%%%%%%%%%%%%%%%%%%%%%%%%%%%%%%%%%%%%%%%%
\textcolor{blue}{\textbf{CIFAR-100 dataset} contains 60000 different object images of size $32\times32$ of 100 different subjects, categorized into 20 super-classes which are considered  as ground-truth~\cite{krizhevsky2009learning}. A random train/test split of 50000/10000 is used in SADSE$_T$. Deep features of dimension 512 are extracted from the second last layer of the baseline algorithm which is Contrastive Clustering (CC) \cite{li2021contrastive} for this dataset. A batch size of 1000 is used in these experiments. The proposed SADSE$_{F}$ has obtained an accuracy of 47.75\% which is 4.85\% better than the baseline CC. It is also better than SPICE, PICA, IIC, and many other methods as shown in Table~\ref{table:cifar20}. Some methods such as  TCL, PCL, SCAN have obtained even better performance which may be attributed to careful fine-tuning after contrastive clustering.}
%%%%%%%%

\begin{table}
\begin{center}
\setlength\tabcolsep{7pt}
\begin{tabular}{|l|c|c|c|c|c|}
\hline %inserts double horizontal lines
{} &  \multicolumn{2}{c}{CIFAR-100} & \multicolumn{2}{c|}{ImageNet-10}\\
\hline
 {\bf Methods} &  {\bf Acc.} & {\bf NMI}&  {\bf Acc.} & {\bf NMI} \\ [0.5ex] % inserts table 
\hline
DAE \cite{vincent2010stacked}&15.1&11.1&30.4&20.6\\
DCGAN \cite{radford2015unsupervised}&15.1&12.0&34.6&22.5\\
DeCNN \cite{zeiler2010deconvolutional}&13.3&9.2&31.3&18.6\\
JULE \cite{yang2016joint}&13.7&10.3&30.0&17.5\\
VAE \cite{kingma2013auto}&15.2&10.8&33.4&19.3\\
ADC \cite{haeusser2019associative}&16.0&-&-&-\\
AE \cite{bengio2006greedy}&16.5&10.00&31.7&21.0\\
DEC \cite{xie2016unsupervised}&18.5&13.6&38.1&28.2\\
DAC \cite{chang2017deep}&23.8&18.5&52.7&39.4\\
IIC \cite{ji2019invariant}&25.7&-&-&-\\
DCCM \cite{wu2019deep}&32.7&28.5&71.0&60.8\\
PICA \cite{huang2020deep}&33.7&31.0&87.0&80.2\\
CC \cite{li2021contrastive}&42.9&43.1&89.3&85.9\\
SPICE \cite{niu2022spice}&46.8&44.8&(96.9)&(92.7)\\
SCAN \cite{van2020scan}&48.3&48.5&-&-\\
PCL~\cite{li2020prototypical}&52.6&52.8&90.7&84.1\\
TCL \cite{li2022twin}&\bf{53.1}&\bf{52.9}&89.5&87.5\\
\hline
SADSE$_{F}$ & 47.75 &45.77& \bf {91.69} &\bf{87.53}\\
\hline 
\end{tabular}
\vspace{-2mm}
\caption{Comparison of the proposed SADSE$_F$  with existing SOTA methods on CIFAR-100 and ImageNet-10 datasets. "()" denotes that pre-training was done using labeled data.} 
\label{table:cifar20}
\end{center}
\end{table}
%%%%%%%%%%%%%%%%%%%%%%%%%%%%%%%%%%%%%%%%%%%%%%
\textcolor{blue}{
\textbf{ImageNet-10 dataset} \cite{chang2017deep} contains 13000 different object images of size $224\times224$ of 10  objects chosen from ILSVRC2012 1K~\cite{deng2009imagenet}. A random train/test split of 10000/3000 is used in SADSE$_T$ while SADSE$_F$ is trained/tested on the full dataset. The same deep features and batch size are employed as in the CIFAR-10 dataset. The proposed SADSE$_{F}$ has obtained an accuracy of 91.69\% outperforming SOTA methods as shown in Table~\ref{table:cifar20}. }
%%%%%%%%%%%%%%%%%%%%%%%%%%%%%%%%%%%%%%%%%%%%%%
\subsection{Generalization to Unseen Data sets}
We evaluate our proposed algorithm in train/test splits as SADSE$_{T}$, and the full dataset used for both training and testing as SADSE$_{F}$. During testing, we don't need to repeat any training step and SADSE$_{T}$ has achieved better test accuracy on all datasets. Many existing methods such as Struct-AE, DSC, and S$^{3}$COMP-C have not reported results on unseen test data.  Overall, we used four measures to demonstrate the generalization of the proposed SADSE algorithm to unseen test data points as shown in Table \ref{table:genscale}. In the ORL dataset, the test/train split is 80/320. Due to the very small training data size, SADSE$_T$ performance is less than SADSE$_F$. While for the other datasets, its performance is the same or better than the full training dataset.

\subsection{Scalabilty to Varying Training Data Size}
Due to batch-based training, both SADSE$_F$ and SADSE$_T$ are scalable to larger datasets. To demonstrate this, we have used an increase of 10000 data points starting from 10000/60000 train/test splits on MNIST dataset. We observed an increase in the accuracy of SADSE$_F$ as training data points are increased, and when testing the same models for SADSE$_T$ the accuracy on unseen test data splits also increases accordingly (Fig. \ref{fig:ablationmnist}a). We observed an increase in the execution time of the SADSE$_F$ as the training dataset becomes larger, and also for SADSE$_T$ when tested on the unseen same number of data points (Fig. \ref{fig:ablationmnist}b). The memory consumption of SADSE$_F$ increases linearly with an increasing dataset as the model gets trained, and also for SADSE$_T$ when tested on the same number of unseen data points (Fig. \ref{fig:ablationmnist}c). The scalability of our proposed method is exhibited when the accuracy of SADSE$_T$ reaches 95\% even when the training data set is only 10000 and all remaining data is used for testing. 
%For all datasets, we have also provided a comparison of time and memory of $\mathcal{S}_b$ using \eqref{eq:eq22} vs. a batch size of full dataset as proposed by many SOTA methods (Fig.~\ref{fig:time_mem_for_H}).
In Table \ref{table:sota_ablation} we have also compared execution time (minutes) and memory consumption (GB)  of SADSE$_F$ with SOTA methods. Due to batch-based training our proposed method, has consumed fewer memory resources and smaller execution time.  
%%%%%%%%%%%%%%%%%%%%%%%%%%%%%%%%%%%%%%%%%%%%%%
%\begin{figure}[h]
%\begin{center}
%{\includegraphics[width=0.50\linewidth]{latex/figures/timeusageofSbondatasets.pdf}\includegraphics[width=0.50\linewidth]{figures/memusageofSbondatasets.pdf}}
%\end{center}
%\vspace{-3mm}
%\caption{ Time \& memory usage of the proposed self expression matrix  computation \mathcall{S}$_b$ batch-wise vs. full data.}
%\label{fig:time_mem_for_H}
%\end{figure}
%%%%%%%%%%%%%%%%%%%%%%%%%%%%%%%%%%%%%%%%%%%%%
\begin{table}
\scriptsize
\begin{center}
\setlength\tabcolsep{0.2pt}
\begin{tabularx}{0.48\textwidth}{|X|X|X|X|X|X|X|X|X|}
\hline
~\bf{Datasets~}& \multicolumn{2}{c|}{\bf DSC}
&\multicolumn{2}{c|}{\bf Struct-AE}
& \multicolumn{2}{c|}{\bf DLRSC}
&\multicolumn{2}{c|}{\bf SADSE$_F$}\\
\cline{2-9}
~&Time&Mem&Time&Mem&Time&Mem&Time&Mem\\
\hline 
~EYaleB&201.0&32.1&175.2&10.66&570.4&0.62&107.1&2.18\\
\hline
~Coil-100~&1653.8&50.6&-&-&3307.6&3.25&458.5&3.58\\
\hline
%MNIST&-&-&-&-&2775.4&13.05&-&-&-&-&13023.2&3.42\\
%\hline
%ORL&1040.08&29.6&-&-&-&-&-&-&&&720.67&2.22\\
%\hline 
\end{tabularx}
\vspace{-2mm}
\caption{Time (min.) and memory (GB) comparison of different SOTA methods with SADSE.} 
\label{table:sota_ablation} 
\end{center}
\end{table}
%\footnotesize{$^{*}$ Time measured for S$^3$COMP-C is in Matlab while other times are measured in python.}
%%%%%%%%%%%%%%%%%%%%%%%%%%%%%%%%%%%%%%%%%%%%%%%%
\subsection{Ablation Study}
To evaluate the contribution of each loss term in SADSE algorithm, we have performed a detailed ablation study. We used different loss terms' combinations to train our proposed model and report the accuracy of SADSE$_F$ in Table ~\ref{table:ablation_on_loss_func}.  We observe each loss term has contributed an increase in the accuracy, and overall loss term $\mathcal{L}_{H}$ provides the best accuracy.

\textbf{Ablation on learning of self-expressive matrix:} Many existing methods compute self-expressive matrix $\mathcal {H}_s$ using Lasso for sparse subspace clustering \cite{Peng23, Elhamifar17}. We implemented the same in a batch fashion to compute $\mathcal {H}_s$ as follows: 
\begin{equation}\label{eq:eq18}
    \mathop {min}_{\mathcal {H}_s}||X_b-X_{b}\mathcal {H}_s||_{F}^{2}+\lambda||\mathcal{H}_s||_{1} \\~~~ s.t. ~ \mathcal{H}_s(i,i)=0,
\end{equation}
and compared it with self-attention-based $\mathcal{S}_b$.
Table \ref{table:ablation_on_H} shows accuracy results on all datasets when $\mathcal{S}_b$ in \eqref{eq:eq19} is replaced by $\mathcal{H}_s$. 
%%%%%%%%%%%%%%%%%%%%%%%%%%%%
\begin{table}
\scriptsize
\begin{center}
\setlength\tabcolsep{0.8pt}
%\resizebox{\columnwidth}{!}{%
\begin{tabularx}{0.47\textwidth}{|X|X|X|X|X|}
\hline
~\bf{Data sets}&~~\bf{Eyaleb}&~~\bf{Coil-100}&~~\bf{MNIST}&~~\bf{ORL}\\
\hline 
$\mathcal{L}_{R}+\mathcal{L}_{S}$ &~~99.95&~~83.59&~~95.61&~~88.75\\
$\mathcal{L}_{o}$&~~99.95&~~84.75&~~96.58&~~89.00\\
$\mathcal{L}_{H}$&~~99.95&~~84.95&~~97.35&~~90.75\\
\hline 
\end{tabularx}
\caption{Ablation study of the proposed SADSE$_F$ algorithm in terms of classification accuracy. The addition of each loss term has caused an increase in the accuracy of the SADSE$_F$ algorithm.} 
\label{table:ablation_on_loss_func} 
\end{center}
\vspace{-5mm}
\end{table}
%%%%%%%%%%%%%%%%%%%%%%%%%%%%%%%%%%%%%
\begin{table}
\scriptsize
\begin{center}
\setlength\tabcolsep{0.8pt}
%\resizebox{\columnwidth}{!}{%
\begin{tabularx}{0.47\textwidth}{|X|X|X|}
\hline
~\bf{Datasets}&~~$\mathbf \mathcal{S}_b$&~~$\mathbf \mathcal{H}_s$\\
\hline 
~~EyaleB&~~99.95&~~99.95\\
~~MNIST&~~97.35&~~97.01\\
~~Coil-100 &~~84.95&~~82.26\\
~~ORL &~~90.75&~~89.25\\
~~CIFAR-100 &~~47.75&~~46.70\\
~~ImageNet-10 &~~91.69&~~92.25\\
%~~GTSRB &~~&~~\\
\hline 
\end{tabularx}
\vspace{-2mm}
\caption{Accuracy comparison using $\mathcal{S}_b$ and $\mathcal{H}_s$ on different datasets.} 
\label{table:ablation_on_H} 
\end{center}
\end{table}
%%%%%%%%%%%%%%%%%%%%%%%%%%%%%%%%%
\noindent{\textbf{Hyper Parameters Tuning:}}
In order to select the best values of  $\lambda_i$, $i=\{1,2,3,4\}$ we have performed experiments on MNIST dataset, and accuracy was observed. We observe that for $\lambda_{1}$=$\lambda_{2}$=$\lambda_{3}$=$\lambda_{4}$=1, the accuracy of SADSE$_F$ is 96.23\%, and with $\lambda_{1}$=$\lambda_{2}$=$\lambda_{3}$=$\lambda_{4}$=0.02, the accuracy is 96.05\%. When we reduce $\lambda_1$ and $\lambda_{3}$ ten times smaller than the others ($\lambda_{1}$=$\lambda_{3}$=0.002, $\lambda_{2}$=$\lambda_{4}$=0.02) the accuracy increases to 97.35\%. So for all datasets, we used the values of hyperparameters as $\lambda_{1}$=$\lambda_{3}$=0.002, and $\lambda_{2}$=$\lambda_{4}$=0.02. Though, further fine-tuning may increase the accuracy of the proposed algorithm.

\textcolor{blue}{Performance variation is observed by varying  $k$=1-6 in KNN on EyaleB dataset and the same performance is observed for 3-6 neighbors. Therefore, for all datasets, $k=3$ is used though  fine-tuning  may 
have further improved the results.}
%%%%%%%%%%%%%%%%%%%%%%%%%%%%%%%%%%%%%%%%%%%%%%%%%
\section{Conclusion}\label{conclusion}
A structure-aware deep spectral embedding (SADSE) algorithm is proposed to learn the spectral representation of input data spanning non-linear manifolds. The proposed SADSE algorithm is based on deep neural networks which are trained by using direct supervision of spectral embedding while preserving the input data structure. The trained network simulates a spectral clustering algorithm including the eigenvector computation of the Laplacian matrix. Therefore the learned representations need not be subjected once again to the traditional spectral clustering as often done by the existing SOTA methods. Thus the learned representations are directly clustered using a linear clustering method such as k-means. To make the learned spectral representation structure aware, a self-expression matrix is batch-wise computed on the input data. 
To this end, a self-attention-based global structure encoding technique is proposed using deep neural networks. The learned self-expression matrix is made sparse by using a nearest-neighbor-based approach. The SADSE algorithm is made scalable to larger datasets by applying loss terms in a batch-wise fashion. Our trained network can also estimate spectral representation for unseen data points coming from distributions similar to the training data. Experiments are performed on four publicly available datasets and compared with existing SOTA methods. Our experiments demonstrate the excellent performance of the proposed SADSE algorithm compared to the existing methods.

%The subspace structure of the input data is encoded by using attention based self-expression learning.
%In the current work, we propose . For this purpose we use two  a query-net and a key-net. These networks are learned by minimizing elastic net constrained self-expression loss. 

%We explicitly preserve self expression based data structure in latent space \& produce a subspace structure aware spectral embedding, which is then linearly clustered to separable subspaces. 

%We separately learn \& evaluate a batch based self expressive matrix to be used in the training of SADSE that learns a mapping from non-linear high dimensional manifolds to low dimensional linear subspaces. 

%%%%%%%%%%%%%%%%%%%%%%%%%%%%%%%%%%%%%%%%
%{\small
%\bibliographystyle{ieee_fullname}
%\bibliography{egbib}
%}

\begin{IEEEbiography}
	[{\includegraphics[width=1in,height=1in,clip,keepaspectratio]{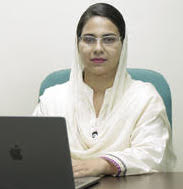}}
	]
	{Hira Yaseen} is a Ph.D. fellow in the Center for Robot Vision, Department of Computer Science, Information Technology University, Lahore, Pakistan. Previously she did her BS in Computer Engineering from the University of Engineering and Technology Lahore and her Master's in Computer Science from Comsats University Islamabad. Her research areas include Computer Vision and Machine Learning. During her Ph.D., she is working on unsupervised representation learning, data clustering, and object classification.
\end{IEEEbiography}

\begin{IEEEbiography}
	[{\includegraphics[width=1.1in,height=1.1in,clip,keepaspectratio]{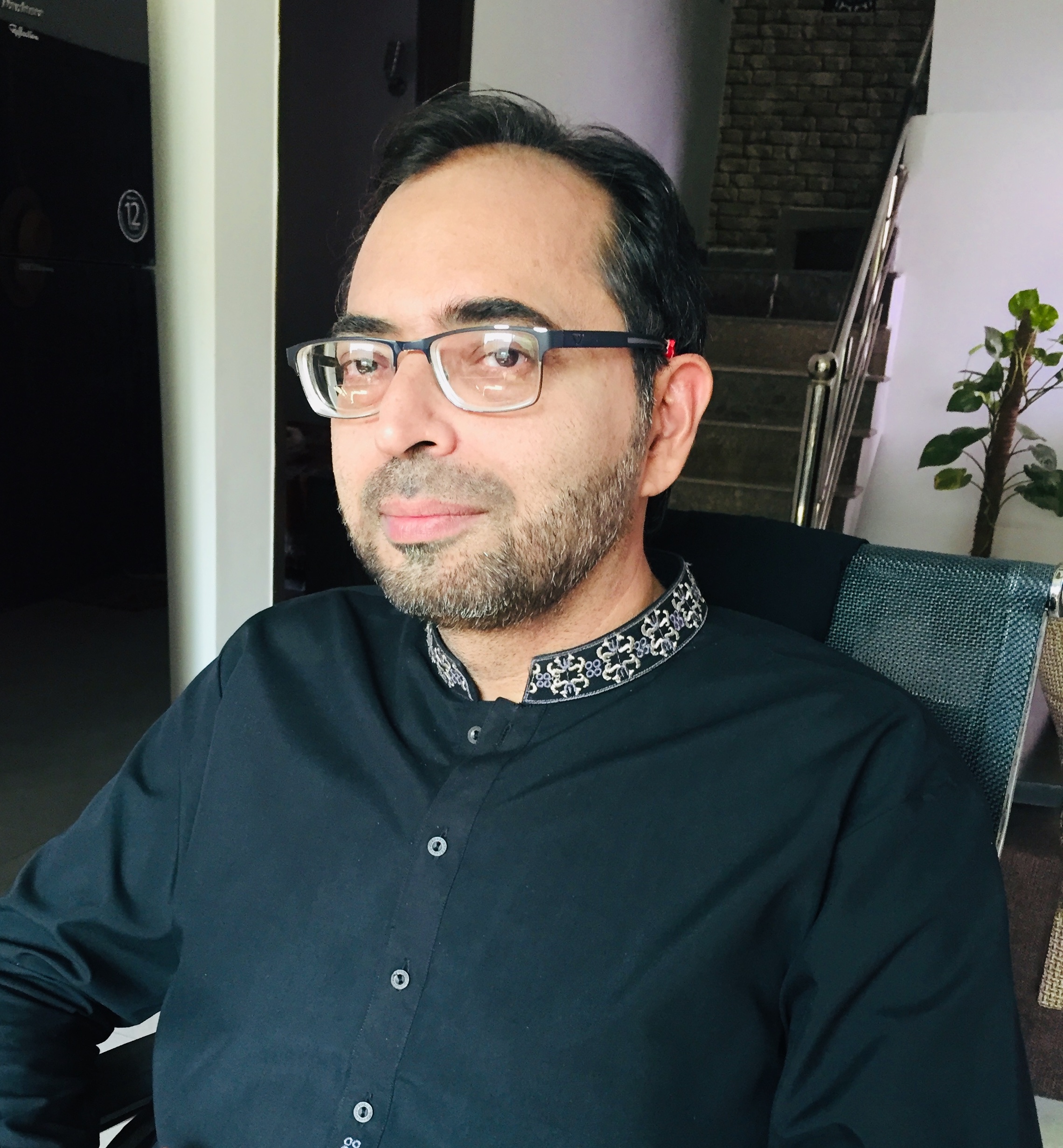}}]{Arif Mahmood} is a Professor in the Computer Science Department, and Dean Faculty of Sciences at the Information Technology University and also Director of the Center for Robot Vision. His current research directions in Computer Vision are person pose detection and segmentation, crowd counting and flow detection, background-foreground modeling in complex scenes, object detection, human-object interaction detection, and abnormal events detection. He is also actively working in diverse Machine Learning applications including  cancer grading and prognostication using histology images, predictive auto-scaling of services hosted on the cloud and the fog infrastructures, and environmental monitoring using remote sensing. He has also worked as a Research Assistant Professor with the School of Mathematics and Statistics, University of Western Australia (UWA) where he worked on Complex Network Analysis. Before that, he was a Research Assistant Professor at the School of Computer Science and Software Engineering, UWA, and performed research on face recognition, object classification, and action recognition. 
\end{IEEEbiography}

\end{document}